\documentclass[conference]{IEEEtran}
\pdfoutput=1
\IEEEoverridecommandlockouts

\usepackage{amsmath,amsfonts,bm}

\newcommand{\figleft}{{\em (Left)}}

\newcommand{\figright}{{\em (Right)}}
\newcommand{\figtop}{{\em (Top)}}
\newcommand{\figbottom}{{\em (Bottom)}}

\def\eqref#1{equation~\ref{#1}}

\def\1{\bm{1}}

\def\eps{{\varepsilon}}

\def\ry{{\textnormal{y}}}

\def\rvx{{\mathbf{x}}}

\def\vone{{\bm{1}}}

\def\vtheta{{\bm{\theta}}}

\def\vb{{\bm{b}}}

\def\vx{{\bm{x}}}
\def\vy{{\bm{y}}}
\def\vz{{\bm{z}}}

\def\vdelta{{\bm{\delta}}}

\def\mB{{\bm{B}}}

\def\mK{{\bm{K}}}

\def\mM{{\bm{M}}}

\def\mQ{{\bm{Q}}}

\def\mV{{\bm{V}}}
\def\mW{{\bm{W}}}
\def\mX{{\bm{X}}}

\DeclareMathAlphabet{\mathsfit}{\encodingdefault}{\sfdefault}{m}{sl}
\SetMathAlphabet{\mathsfit}{bold}{\encodingdefault}{\sfdefault}{bx}{n}

\def\sS{{\mathbb{S}}}

\newcommand{\pdata}{p_{\rm{data}}}

\newcommand{\Ls}{\mathcal{L}}

\newcommand{\Cov}{\mathrm{Cov}}

\newcommand{\normltwo}{L^2}

\DeclareMathOperator*{\argmin}{arg\,min}

\DeclareMathOperator{\layn}{LN}
\DeclareMathOperator{\mlp}{MLP}
\DeclareMathOperator{\msa}{MSA}
\DeclareMathOperator{\diag}{diag}

\newcommand{\oldtext}[1]{{}}

\newcommand{\cmark}{\color{black!75}\ding{51}}%
\newcommand{\xmark}{\color{black!50}\ding{55}}%

\newcommand{\vizwidht}{0.7\textwidth}%

\renewcommand{\smallskip}{\vspace{2pt}}

\usepackage{amsmath,amssymb,amsfonts}
\usepackage{algorithmic}
\usepackage{textcomp}
\usepackage[bibstyle=ieee,citestyle=numeric-comp,mincitenames=1,maxcitenames=2]{biblatex} %
\bibliography{references.bib}

\usepackage[hidelinks]{hyperref}
\usepackage{url}
\usepackage{nicefrac}       %
\usepackage{microtype}      %
\usepackage{booktabs}       %
\usepackage[accsupp]{axessibility}  %

\usepackage{color, colortbl}
\usepackage{multirow}
\usepackage{graphicx}
\usepackage{multirow}
\usepackage{dsfont}
\usepackage{wrapfig}
\usepackage{xcolor}
\usepackage{pifont}
\usepackage{enumitem}
\usepackage{subcaption}
\usepackage[nameinlink,capitalize]{cleveref}
\usepackage{lipsum}
\usepackage[normalem]{ulem}
\usepackage{threeparttable}
\usepackage{bm}
\usepackage[most]{tcolorbox}
\usepackage{tabularx}

\newtcbtheorem{Summary}{}{enhanced,
  coltitle=black,
  top=0.14in,
  attach boxed title to top left=
  {xshift=0em,yshift=-\tcboxedtitleheight/2},
  boxed title style={size=small,colback=blue!10}
}{summary}

\newcommand{\citep}{\cite}
\newcommand{\citet}{\textcite}

\def\BibTeX{{\rm B\kern-.05em{\sc i\kern-.025em b}\kern-.08em
    T\kern-.1667em\lower.7ex\hbox{E}\kern-.125emX}}

\begin{document}

\title{A \textit{Light} Recipe to Train Robust Vision Transformers\\
\thanks{$^{\dagger}$Work done as a Master's student at the IC Department at EPFL.}
}

\author{\IEEEauthorblockN{Edoardo Debenedetti$^{\dagger}$}
\IEEEauthorblockA{\textit{Department of Computer Science} \\
\textit{ETH Zurich}\\
Z{\"u}rich, Switzerland \\
\texttt{edoardo.debenedetti@inf.ethz.ch}}
\and
\IEEEauthorblockN{Vikash Sehwag, Prateek Mittal}
\IEEEauthorblockA{\textit{Department of Electrical and Computer Engineering} \\
\textit{Princeton University}\\
Princeton, United States \\
\texttt{\{vvikash,pmittal\}@princeton.edu}}
}

\maketitle

\begin{abstract}
In this paper, we ask whether Vision Transformers (ViTs) can serve as an underlying architecture for improving the adversarial robustness of machine learning models against evasion attacks. While earlier works have focused on improving Convolutional Neural Networks, we show that also ViTs are highly suitable for adversarial training to achieve competitive performance. We achieve this objective using a custom adversarial training recipe, discovered using rigorous ablation studies on a subset of the ImageNet dataset. The canonical training recipe for ViTs recommends strong data augmentation, in part to compensate for the lack of vision inductive bias of attention modules, when compared to convolutions. We show that this recipe achieves suboptimal performance when used for adversarial training. In contrast, we find that omitting all heavy data augmentation, and adding some additional bag-of-tricks ($\varepsilon$-warmup and larger weight decay), significantly boosts the performance of robust ViTs. We show that our recipe generalizes to different classes of ViT architectures and large-scale models on full ImageNet-1k. Additionally, investigating the reasons for the robustness of our models, we show that it is easier to generate strong attacks during training when using our recipe and that this leads to better robustness at test time. Finally, we further study one consequence of adversarial training by proposing a way to quantify the semantic nature of adversarial perturbations and highlight its correlation with the robustness of the model. Overall, we recommend that the community should avoid translating the canonical training recipes in ViTs to robust training and rethink common training choices in the context of adversarial training. We share the code for our experiments at {\color{blue}\url{https://github.com/dedeswim/vits-robustness-torch}}.
\end{abstract}


\section{Introduction}

\begin{figure}[ht]
    \centering
    \includegraphics[width=\linewidth]{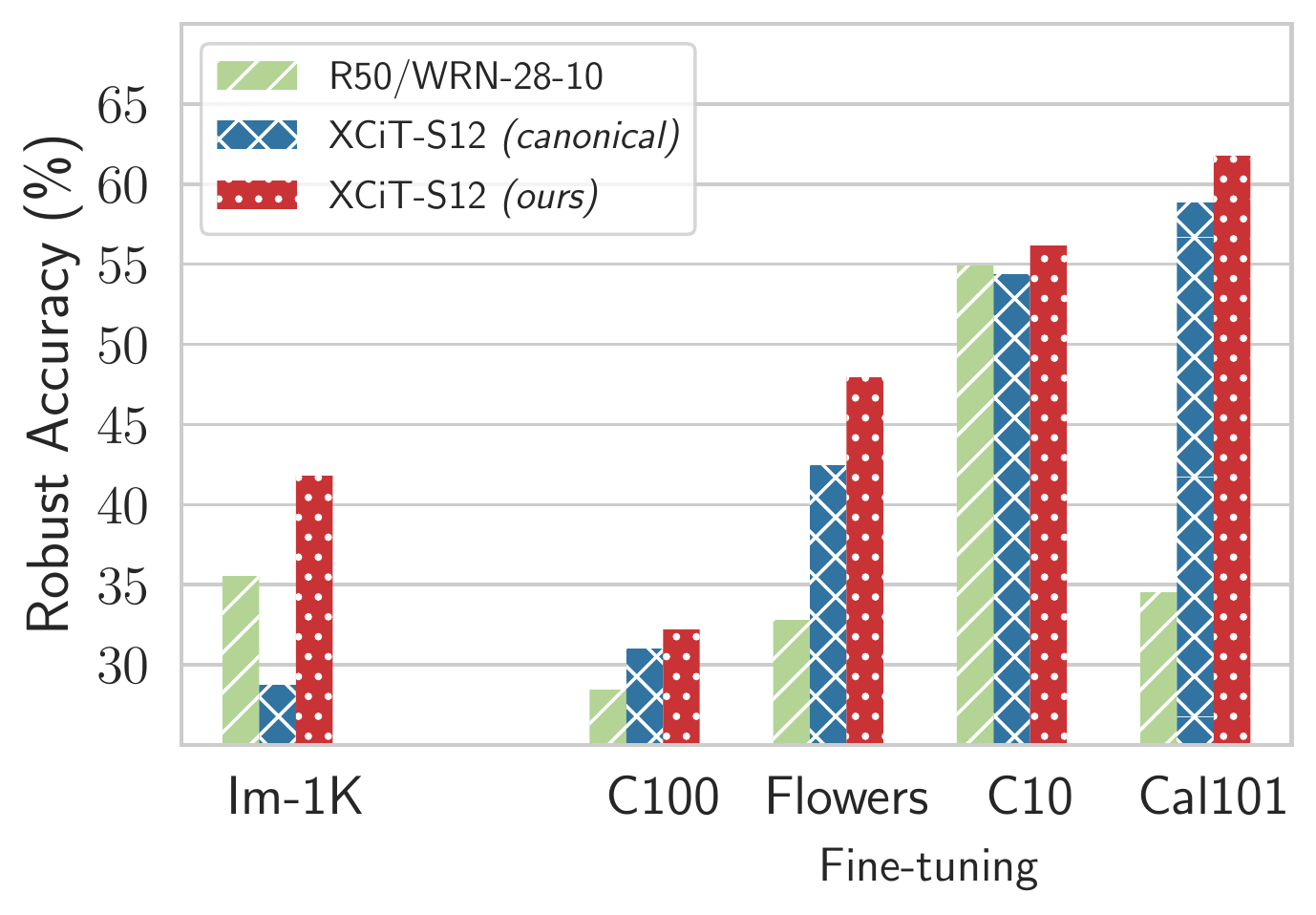}
    \caption{\textbf{A \textit{light} recipe is better!} Comparison of our proposed recipe with light data augmentation and the canonical one for ViTs (XCiT-S12 in this experiment) in pre-training on ImageNet-1k and finetuning on four different datasets. Our recipe boosts robust accuracy by $13.1\%$ on ImageNet-1k and up to $5.5\%$ on downstream finetuning tasks. For further comparison, we also include ResNet-50 (with smooth activation function~\cite{xie2020smooth, bai2021transformers} -- GELU~\cite{hendrycks2016gaussian}). Since a WideResNet-28-10 performs better for low-resolution datasets, we use it as a baseline for C10 and C100. Abbreviations for datasets are 1) \textit{Im-1k}: ImageNet-1k, 2) \textit{C100}: CIFAR-100, 3) \textit{Flowers}: Oxford-flowers, 4) \textit{C10}: CIFAR-10, and 5) \textit{Ca101}: Caltech-101.}
    \label{fig:summary_intro}
\end{figure}

\noindent Adversarial training~\cite{madry2017towards}, which has emerged as one of the most successful defenses against evasion attacks targeting machine learning models~\cite{biggio2013evasion,szegedy2013intriguing}, consists of training against worst-case inputs to make networks robust. However, it incurs a large generalization gap~\cite{tsipras2018robustness, lee2020adversarial, schmidt2018adversarially}, and closing this gap is a key challenge. Existing work that tackles this problem can be broadly classified into three categories: 1) improvements in the adversarial training \emph{mechanism}~\cite{madry2017towards,zhang2019theoretically,wang2020improving,wu2020adversarial}, 2) modifications to the \emph{training data}~\cite{rebuffi2021data,gowal2021improving,sehwag2021robust}, and 3) modifications of the \emph{neural network architecture} for robust training~\cite{xie2020smooth, dai2022parameterizing}. Our work focuses on the latter category, by improving robustness with new architectures while improving the training recipe. %

\smallskip
\noindent \textbf{Vision Transformers for adversarial training.} Several earlier works have innovated on activation functions~\cite{xie2020smooth,dai2022parameterizing,bai2021transformers} for CNNs or their structure~\cite{huang2021exploring} to improve robustness. While such advancements in CNNs are helpful and CNNs are the \emph{de facto} standard architecture for adversarial training, we show that a drastic boost in adversarial robustness can be achieved by switching the architecture class itself, i.e., by using Vision Transformers (ViTs)~\cite{dosovitskiy2021image}. In fact, even non-adversarially trained ViTs show some signs of higher robustness than conventional CNNs when considering non-adversarial perturbations~\cite{paul2021vision,bhojanapalli2021understanding,bai2021transformers,aldahdooh2021reveal}, which further encourages their use in robust training. Inspired by these features shown by ViTs, prior work~\cite{bai2021transformers,wu2022towards,shao2021adversarial} has attempted to train ViTs with adversarial training but failed to replace CNNs as the dominant architecture on robustness benchmarks~\cite{croce2020robustbench}. Since ViTs are competitive with CNNs in standard training~\cite{khan2021transformersSurvey,pwcImageNet}, it is imperative to ask whether their suboptimal performance in adversarial training is fundamental or simply an artifact of sub-par training recipes. With a systematic investigation, we identify that the latter is true and propose a simple, yet highly effective recipe to boost the performance of ViTs in adversarial training.

\smallskip \noindent \textbf{Canonical ViTs training recipes are suboptimal for adversarial training.}
Due to the lack of strong inductive bias in ViTs, they need custom training recipes~\citep{steiner2021train}, which include heavy data augmentation, to achieve optimal performance. We observe that using these canonical recipes with adversarial training leads to sub-optimal performance, even suggesting that state-of-the-art ViTs~\cite{elnouby2021xcit} fail to significantly outperform conventional CNN architectures, such as ResNets~\cite{he2015deep}. We show that optimizing the training recipe boosts the performance of ViTs from suboptimal to \textit{state-of-the-art} (\cref{fig:summary_intro}).

\smallskip \noindent \textbf{Improved training recipe (weaker data augmentation is better!).} Instead of just taking the canonical training recipe used by DeiT~\citep{touvron2021training} and successive works~\citep{touvron2021going,elnouby2021xcit,yu2021metaformer,graham2021levit}, we first analyze some ViT variations and architectural components that could make ViTs more suitable for adversarial training. We then identify a set of important parameters that have a fundamental role in adversarial training, such as adversarial training warm-up, data augmentations, and weight decay. We go beyond the canonical training recipes commonly used for ViT-like models by doing a thorough search for the optimal values of these parameters: we observe that the optimal choices for non-adversarial training drastically differ from those for adversarial training. In fact, we find that while the use of strong data augmentation is recommended for standard training~\citep{steiner2021train,touvron2021training}, this is detrimental to adversarial training: not only does using light data augmentation improve the robustness of ViTs, but it does so without sacrificing the clean accuracy. Our recommendation to use weaker data augmentation is also \textit{surprising} since earlier works have argued for stronger data augmentation in adversarial training~\cite{bai2021transformers}, albeit with a warm-up in the augmentation intensity, as well as the same weight decay as the one used for standard training.

\smallskip \noindent \textbf{Generalization of our recipe across scales, architectures, and datasets.} 
Improving over the canonical training recipe requires a rigorous ablation study, which has a high computational cost on large-scale datasets and models.\footnote{Doing such ablation on full ImageNet-1k with XCiT-S would take 2632 TPUv3 hours (i.e., 179 TPUv3 days).} To circumvent this challenge, we optimize our recipe with a subset of the ImageNet-1k dataset and smaller models. However, for widespread usage, it is imperative that the benefits of our recipe generalize across diverse scenarios. We first show that the proposed recipe outperforms the canonical one at scale, i.e., on the full ImageNet-1k dataset. Next, we show that not only it performs better across the XCiT transformer class~\cite{elnouby2021xcit}, but also for other transformers (such as DeiT~\cite{touvron2021training} and PoolFormer~\cite{yu2021metaformer}), as well as for modern CNNs (ConvNeXt~\cite{liu2022convnet}). Finally, we show that the gains of our recipe in pre-training on ImageNet-1k also directly transfer when finetuning on smaller datasets of both high- and low- resolution.

\smallskip
\noindent \textbf{Delving deeper into adversarial robustness of ViTs.} We conduct two analyses to better understand why our recipe brings an improvement: attack effectiveness and semantic nature quantification. Since large-scale adversarial training uses only few-step attacks, its performance depends on the strength of adversarial examples generated with few-step attacks. We uncover that, throughout the whole training duration, few-step attacks are more effective for a ViT trained with our recipe than for a ViT trained with the canonical recipe, as well as more effective than a conventional CNN architecture. This results in a model which is overall more robust to strong attacks at test time. Next, we propose a new way to quantify the semantic nature of adversarial perturbations~\citep{engstrom2019adversarial} for robust ViTs. We show, with a quantitative method, that the adversarial perturbations targeting XCiT-S12 have more semantic features than those targeting a GELU ResNet-50.

\smallskip 
\noindent \textbf{Key contributions.} We make the following key contributions:
\begin{itemize}[leftmargin=10pt]
\item Through a rigorous ablation study, we uncover a light yet effective adversarial training recipe for ViTs. In particular, we find that the use of weak data augmentation, in contrast to strong augmentation in the canonical recipe, achieves \textit{state-of-the-art} performance on the ImageNet-1k dataset (compared with the other models on RobustBench~\cite{croce2020robustbench}), with models having up to $47.60\%$ AutoAttack accuracy and $73.76\%$ clean accuracy.
\item We further show that our proposed recipe generalizes across different scales of datasets and models, and different classes of ViT architectures.
\item We demonstrate that the advantage of our recipe in pre-training also leads to benefits when finetuning on downstream datasets. Across four low- and high-resolution datasets, our recipe achieves up to $5.5\%$ higher robust accuracy than the canonical recipe.
    \item We identify that the high robustness of ViTs is also related to the effectiveness of adversarial attacks on them. Simultaneously, we quantify that adversarial perturbations targeting robust ViTs have semantic characteristics.  
\end{itemize}

\smallskip 
\noindent \textbf{Paper Outline.} We provide a brief overview of adversarial training and the ViT architecture in \cref{sec:background}. In \cref{sec:experiments}, we first identify the limitations of the canonical recipe and then conduct an ablation study for the proposed training recipe. In \cref{sec:val-at-scale}, we first show the advantage of our recipe on the full-scale ImageNet-1k dataset across different architectures. We later demonstrate its downstream benefits in finetuning. In \cref{sec:understanding}, we examine why ViTs are highly successful in adversarial training. In \cref{sec:related-work}, we present the related works and conclude the paper with a discussion in \cref{sec:conclusion}.

\smallskip
\noindent \textbf{Open-sourced artifacts.} Finally, we share the code on GitHub\footnote{\url{https://github.com/dedeswim/vits-robustness-torch}} to enable the community both to reproduce our results and fine-tune the models on further datasets. In the same repository, we also share the checkpoints of our models for five datasets, numerous architectures, and perturbations to enable other researchers to fine-tune the models, as well as run further analyses.
\section{Background}\label{sec:background}
Adversarial training~\cite{madry2017towards} has shown to be an effective and reliable method to defend against adversarial examples, and most of the subsequent work about defenses against adversarial examples is based on it. On the other hand, ViTs are an emerging class of architectures that achieve competitive performance on standard computer vision benchmarks~\cite{khan2021transformersSurvey}. In this section, we give an overview of both.

\subsection{Overview of adversarial training}
Adversarial training~\cite{madry2017towards} is one of the most successful defenses against adversarial examples.
It consists of generating adversarial examples at each step and training the model using the generated perturbed data instead of the original, clean data. Given a model with parameters $\vtheta$, input data $\rvx$ with label $\ry$ sampled from a distribution $\pdata$, a set of allowed inputs $\sS$ and a loss $\Ls$, adversarial training formally consists of optimizing the following $\min$-$\max$ problem:
\begin{equation}\label{eq:adv-train}
	\hat{\vtheta} = \argmin_\vtheta \mathbb{E}_{(\rvx, \ry) \sim \pdata}  \left[ \max_{\vdelta \in \sS} \Ls(\rvx + \vdelta, \ry; \vtheta) \right].
\end{equation}
\noindent Several further techniques have been proposed to improve adversarial training and the trade-off~\cite{zhang2019theoretically} between robustness and accuracy. Some of them include the optimization of a different objective~\cite{zhang2019theoretically, rade2021helperbased}, early stopping to avoid the so-called robust overfitting~\cite{rice2020overfitting}, tuned data augmentation~\cite{rebuffi2021data}, weight averaging~\cite{rebuffi2021data} etc.

\smallskip
\noindent \textbf{Metrics in adversarial training.} It is common to use the following two metrics to measure performance in adversarial training: a) \textit{Clean accuracy}: the accuracy of the model evaluated on the original data. b) \textit{Robust accuracy.}: the accuracy of the model on a dataset of adversarial examples generated from the original data using an attack. To avoid overestimating the robustness of our networks, it is critical to evaluate the models using a strong attack when generating adversarial examples. We employ AutoAttack~\cite{croce2020reliable} for the robust accuracy evaluation. AutoAttack is considered the de-facto standardized method to assess the adversarial robustness of classification models. It is an ensemble of four different parameter-free attacks, three white- and one black-box. We provide more details about the evaluation procedure of our experiments in~\cref{sec:experiments}.

\subsection{Overview of Vision Transformers}

\begin{figure}
    \centering
    \includegraphics[width=\linewidth]{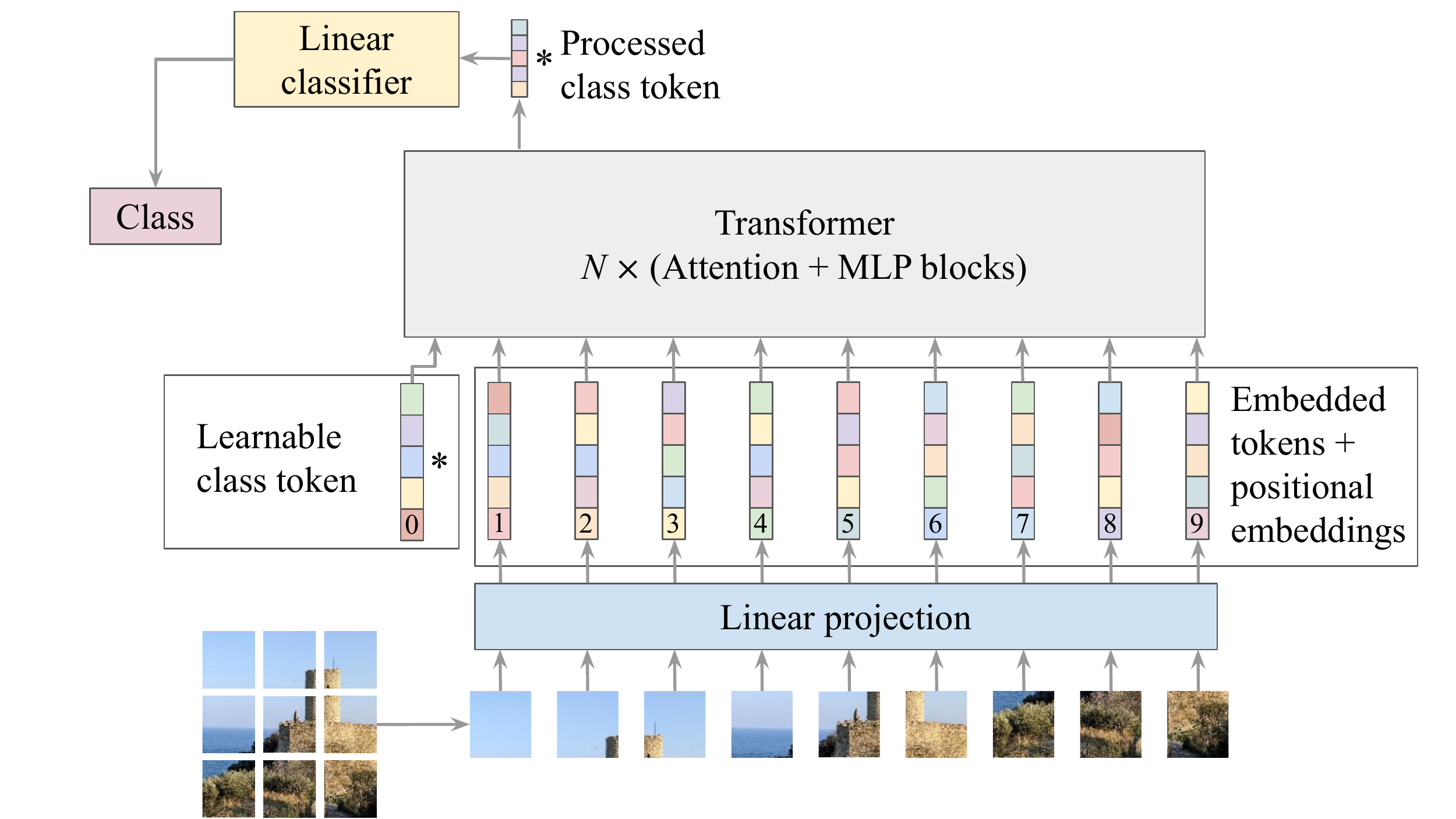}
    \caption{\textbf{Overview of the transformer architecture.} The different phases of a vision transformer model: 1) patchification 2) addition of the positional encodings and of the \texttt{[cls]} token 3) processing through $N$ transformer blocks 4) classification. Further details about ViTs and some variants are in \cref{sec:vision-transformer}. Image reproduced from \textcite{dosovitskiy2021image}.}
    \label{fig:vit}
\end{figure}

The Transformer architecture~\cite{vaswani2017attention} is an architecture for sequence transduction and Natural Language Processing (NLP) tasks (e.g., machine translation) based on the attention mechanism~\cite{bahdanau2014neural}. This architecture includes a series of so-called \emph{Multi-Head Attention} layers, each followed by a \emph{Multi-Layer Perceptron} block. Every layer has a residual connection. This architecture takes as input a series of words, which are tokenized and embedded into vectors of size $d_\text{model}$. The Transformer architecture can also be easily adapted for computer vision tasks~\cite{dosovitskiy2021image}. The resulting architecture is called \emph{Vision Transformer} (ViT). In particular, it is possible to divide the input images into non-overlapping patches, which are embedded into tokens, and then fed to the transformer. We give additional in-depth details about the various components of this architecture, as well as more information about the ViT variants used in this work in \cref{sec:vision-transformer}.

\section{Finding an effective adversarial training recipe for vision transformers}\label{sec:experiments}

We now aim at finding an effective training recipe to leverage the potential that ViTs have shown for standard training. We first highlight the limitations of the canonical standard training recipes which use strong data augmentation. We then conduct an ablation study of major design choices: 1) a warm-up for the perturbation budget, 2) data augmentation policy, and 3) weight decay. As a result, we show the factor which influences performance the most: surprisingly, strong data augmentation leads to sub-par performance.

\smallskip \noindent \textbf{Setup.} We validate the ablations in this section with a simplified setup for the sake of efficiency: we train thirty-two models on a subset of 100 random classes of ImageNet-1k and validate them using APGD-CE~\citep{croce2020reliable}, the first attack of the well-established AutoAttack~\cite{croce2020reliable}. However, in \cref{sec:val-at-scale} we will validate the findings of this section on the full ImageNet-1k dataset~\cite{deng2009imagenet}, using the full ensemble of AutoAttack. Throughout the paper, we mainly focus on untargeted $\ell_\infty$ attacks, and for ImageNet-100 and ImageNet-1k on perturbations of magnitude $\nicefrac{4}{255}$, as this is the most studied scenario so far~\cite{croce2020robustbench} (with the exception of \cref{sec:im-validation}, where we also test our recipe on the $\ell_2$ threat model with $\eps = 3.0$). We evaluate the checkpoint which has the highest FGSM accuracy throughout the training, to prevent robust overfitting~\cite{rice2020overfitting}. Additional details about the training setup and procedure are in \cref{sec:setup}.

\subsection{The canonical training recipe leads to suboptimal performance in adversarial training}

While early attempts with ViTs succeeded only with large-scale pre-training, a lot of effort has been put to achieve good performance without the need of pre-training them on very large datasets. In particular, both~\citet{touvron2021training} and~\citet{steiner2021train} observe that strong data augmentation techniques are needed, i.e., MixUp, CutMix, RandAugment, and Random Erasing, as they help compensate for the lack of a strong vision prior such as convolutions. Similarly, other work on ResNets and CNNs, in general, observed that stronger data augmentation improves the generalization of standard training~\citep{wightman2021resnet,liu2022convnet}.

\smallskip \noindent \textbf{Canonical training recipe.} We refer to \textit{canonical training recipe} as the one used in the original XCiT paper~\citep{elnouby2021xcit}, which, in turn, is borrowed from DeiT's paper~\cite{touvron2021training}, and used for the large majority of ViT variations (e.g., CaiT~\cite{touvron2021going} and PoolFormer~\cite{yu2021metaformer}), and modern CNNs (e.g., ConvNeXt~\cite{liu2022convnet}). We summarize this recipe below. When adversarially training XCiT-S, a ViT variant, if we use the canonical recipe, we observe very poor performance when compared to the equivalent state-of-the-art ResNet-50 with GELU activation function by \textcite{bai2021transformers}. They use a setup that does not differ from the standard setup for ResNets and achieve almost one fourth better robust accuracy ($35.51\%$ vs. $28.70\%$). Given this difference, considering that on standard training XCiT-S performs better than ResNet-50, it is natural to investigate whether a better setup can lead to stronger results. In this section, we rigorously analyze the design space of adversarial training and propose a better training recipe. In \cref{tab:canonical-limitations}, we show a summary of why the canonical recipe is sub-optimal for XCiT-S, with further, rigorous results in \cref{tab:big_ablation}, which we discuss in the next sections.

\begin{table}[h]
    \centering
    \caption[Caption for LOF]{\textbf{Limitation of canonical training recipe.} While borrowing the canonical recipe in adversarial training largely succeeds for CNNs, i.e., ResNet-50, it leads to suboptimal performance in ViTs, i.e., XCiT-S. Directly translating the canonical recipe to adversarial training gives a \textit{false} impression that ViTs are not suitable for adversarial training.}\label{tab:canonical-limitations}
    \resizebox{\linewidth}{!}{
    \begin{tabular}{lcc}
        \toprule
        \multirow{2}{*}{Architecture}   & Standard training          & Adversarial training   \\ \cmidrule{2-3}
                                                                    & Clean accuracy & AutoAttack accuracy                      \\ \midrule
        ResNet-50   & $76.0$ \cite{paszke2017automatic,wightman2021resnet} \footnotemark  & $\bm{35.51}$ \cite{bai2021transformers}                          \\
        XCiT-S                             & $\bm{82.0}$ \cite{elnouby2021xcit} & $28.70$                          \\ 
        \bottomrule
    \end{tabular}}
\end{table}

\footnotetext{For ResNet we show the clean accuracy of the model available in the torchvision library, as reported by~\textcite{wightman2021resnet}. However, this model is trained without heavy augmentation, similarly to the adversarially trained ResNet-50 by \textcite{bai2021transformers}, even though \textcite{wightman2021resnet} show that heavy data augmentation benefits ResNets as well.}

\begin{Summary}[title=\textbf{Canonical training recipe}]{}{}
\begin{itemize}[leftmargin=0pt]
    \item Strong data augmentations (MixUp + CutMix + RandAugment + Random Erasing)
    \item Small weight decay ($0.05$ using AdamW on ImageNet-1k)
\end{itemize}
\end{Summary}

\begin{table*}
    \caption{\textbf{Beyond canonical choices: identifying best adversarial training setup for ViTs.} We analyze choices in architecture, data-augmentation, and optimization setup to identify the best training setup for adversarial training. We use the ImageNet-100 dataset and measure the robust accuracy using the APGD-CE attack. Our analysis uncovers intriguing trends: one is a counter-intuitive phenomenon where \textit{weaker} data augmentations lead to better performance in ViTs with adversarial training.}
    \label{tab:big_ablation}
    \begin{subtable}[t]{0.65\textwidth}
    \footnotesize
	\renewcommand{\arraystretch}{1.2}
	\centering
	\caption{\textbf{Comparison of different ViT-like architectures.} Innovation in ViTs architectures has a relevant impact on adversarial training performance. We observe that the cross-covariance attention-based XCiT architecture achieves significantly better performance than the others.}
	\label{tab:architectures}
    	\begin{tabular}{lccccc} \toprule
    		\multirow{2}{*}{Architecture}      & \multirow{2}{*}{Parameters} & \multirow{2}{*}{GFLOPs} & \multicolumn{2}{c}{Accuracy}       \\ \cmidrule{4-5}
    		                                   &                             &                         & \textit{Clean}  & \textit{APGD-CE} \\ \midrule
    		DeiT-S~\citep{touvron2021training} & 22M                         & 4.61                    & 62.52           & 33.32            \\
    		CaiT-S-12~\citep{touvron2021going} & 25M                         & 4.76                    & 70.20           & 35.84            \\
    		XCiT-N12~\citep{elnouby2021xcit}   & 3M                          & 0.56                    & 48.46           & 30.48             \\
    		XCiT-S12~\citep{elnouby2021xcit}   & 26M                         & 4.81                    & \textbf{85.06}  & \textbf{54.80}   \\ \bottomrule
    	\end{tabular}
    \end{subtable}
    \hfill
    \begin{subtable}[t]{0.33\textwidth}
    	\centering
    	\footnotesize
    	\caption{\textbf{Attack curriculum.} We warm up $\eps$ by linearly increasing it for a fixed number of early epochs. It benefits both clean and robust accuracy. (Arch: XCiT-N12).}\label{tab:eps-warmup}
    	\begin{tabular}{lcc} \toprule
    		\multirow{2}{*}{Epochs} & \multicolumn{2}{c}{Accuracy}                    \\ \cmidrule{2-3}
    		                        & \textit{Clean}               & \textit{APGD-CE} \\ \midrule
    		0                       & 48.46                        & 30.48            \\
    		5                       & 52.04                        & 32.86            \\
    		10                      & 54.62                        & 33.84            \\
    		20                      & 56.10                        & \textbf{34.88}   \\
    		30                      & \textbf{56.12}               & 34.54            \\ \bottomrule
    	\end{tabular}
    \end{subtable}
    \newline
    \vspace*{0.2 cm}
    \newline
    \begin{subtable}[t]{0.65\linewidth}
        \footnotesize
    	\renewcommand{\arraystretch}{1.2}
    	\centering
    	\caption{\textbf{Weak data augmentation is better.} The strategies that perform best are those with just Random Erasing or no heavy augmentation at all. We report the seven best results, and the baseline recipe with all data augmentations, in this table (full results in \cref{tab:data-aug-full} in \cref{sec:add-results}). In all the runs in this table we keep weak data augmentations that are commonly used (random flip and crop, and color jitter). (Arch: XCiT-N12)}\label{tab:data-aug}
    	\begin{tabular}{ccccccc} \toprule
			\multicolumn{4}{c}{Data Augmentation Policy} &                                                                                                                                                           & \multicolumn{2}{c}{Accuracy} \\ \midrule
			MixUp                                        & CutMix & RandAugment                                                 & Random Erasing                                                                     & & \textit{Clean} & \textit{APGD-CE} \\ \midrule
			\xmark                                       & \xmark & \xmark                                                      & \cmark                                                                             & & \textbf{67.28} & \textbf{39.22}   \\
			\xmark                                       & \xmark & \xmark                                                      & \xmark                                                                             & & 66.78          & \textbf{39.22}   \\
			\cmark                                       & \xmark & \xmark                                                      & \xmark                                                                             & & 61.04          & 38.56            \\
			\cmark                                       & \xmark & \xmark                                                      & \cmark                                                                             & & 60.46          & 38.26            \\
			\cmark                                       & \cmark & \xmark                                                      & \xmark                                                                             & & 62.04          & 38.18            \\
			\xmark                                       & \xmark & \cmark                                                      & \xmark                                                                             & & 65.34          & 37.64            \\
			\xmark                                       & \xmark & \cmark                                                      & \cmark                                                                             & & 64.76          & 37.62            \\
			\cmark                                       & \cmark & \cmark                                                      & \cmark                                                                             & & 56.64          & 35.38            \\ \bottomrule
		\end{tabular}
    \end{subtable}
    \hfill
	\begin{subtable}[t]{0.32\linewidth}
    	\footnotesize
    	\renewcommand{\arraystretch}{1.29}
    	\centering
    	\caption{\textbf{Large weight decay helps.} The best results are obtained with $0.5$ weight decay, which is $10\times$ larger than the $0.05$ weight decay used in the canonical recipe. (Arch: XCiT-N12)}\label{tab:weight-decay}
    	\begin{tabular}{lcc} \toprule
    		\multirow{2}{*}{Weight Decay} & \multicolumn{2}{c}{Accuracy}       \\ \cmidrule{2-3}
    		                              & \textit{Clean}  & \textit{APGD-CE} \\ \midrule
    		0                             & 66.44           & 39.02            \\
    		0.001                         & 66.40           & 39.04            \\
    		0.01                          & 66.28           & 38.66            \\
    		0.05                          & 67.16           & 39.30            \\
    		0.1                           & 67.28           & 39.92            \\
    		0.5                           & \textbf{68.78}  & \textbf{42.02}   \\
    		1.0                           & 67.68           & 40.88            \\ \bottomrule
    	\end{tabular}
    \end{subtable}
\end{table*}

\subsection{Architecture choice: Architectural innovations significantly benefit adversarial training}\label{sec:arch-ablation}
Our objective is to understand how innovations in ViTs architecture, post the conception in~\citet{dosovitskiy2021image}, directly benefit adversarial training. In particular, we focus on DeiT~\citep{touvron2021training}, CaiT~\citep{touvron2021going}, and XCiT~\citep{elnouby2021xcit}. These architectures are a natural choice, as each improves upon the latter. CaiT improves over DeiT by introducing \textit{Class Attention} while XCiT further improves CaiT using \textit{Cross-Covariance Attention}. We provide detailed descriptions of each architecture in \cref{sec:transformer-arch}. Our results in \cref{tab:architectures} show that Class Attention in CaiT helps with the fit to adversarial training, and Cross-Covariance Attention boosts, even more, the performance. For this reason, we choose XCiT as the base architecture for our experiments. To speed up the ablation study, we use a smaller variant of XCiT, XCiT-N12.

\subsection{Data-augmentation: Adversarial training of ViTs requires weak augmentation}

The success of ViTs strongly depends on the use of heavy data augmentation and appropriate regularization (\citet{touvron2021training, steiner2021train}), often attributed to their lack of inductive bias for vision tasks. Simultaneously, adversarial training for CNNs also benefits from heavy data augmentation~\citep{rebuffi2021data}, albeit on low-resolution datasets, thus it is natural to start using heavy data augmentation in adversarial training of ViTs. However, we find this choice highly sub-optimal, as \textit{weaker} data augmentation achieves much better performance with adversarial training (\cref{tab:data-aug}). We run a thorough ablation, considering all sixteen combinations of the four key data augmentation policies: CutMix~\citep{yun2019cutmix}, RandAugment~\citep{cubuk2020randaugment}, MixUp~\citep{zhang2017mixup}, and Random Erasing~\citep{zhong2020random}. We give an in-depth description and examples of the various data augmentation techniques in \cref{sec:data-augmentation}. In all the training runs, we always apply basic augmentations such as horizontal flipping, random resize-rescale, and color jitter. We show the top seven setups in \cref{tab:data-aug}, ranked by APGD-CE accuracy (full results in \cref{tab:data-aug-full} in \cref{sec:add-results}). Surprisingly, the augmentation setup that leads to the best results in terms of APGD-CE accuracy is the one with no additional augmentations, apart from the basic ones listed above, together with the one that uses only Random Erasing. These setups improve the robust accuracy by 3.84\% over the canonical strategy of using heavy data augmentation. This phenomenon is likely arising due to the inherent regularization imposed by adversarial training, where strong adversarial perturbations already make the optimization much harder thus leading to better performance without heavy augmentation. Moreover, we note that the models with the best robust accuracy are also the ones with the best clean accuracy, suggesting that using only light data augmentation does not affect clean accuracy. We note that the setup with only Random Erasing has the same robust accuracy as the one with no heavy augmentation but has a slightly larger clean accuracy. Despite this, we choose as a setup for the next experiments the one without Random Erasing, to keep the overall setup as simple as possible. We will validate this choice in \cref{sec:im-validation}.

\subsection{Optimization setup: Tuning attack curriculum and additional regularization brings further improvements}

\smallskip \noindent \textbf{Epsilon warm-up.}
\citet{bai2021transformers} observe that adversarially training a DeiT on the full ImageNet-1k dataset would fail using the same setup as the one in the DeiT paper. For this reason, we attempt training an XCiT-N12 to see if the training succeeds. Even though the training run succeeds, the model struggles in the first few epochs. A possible solution could be to use the following attack curriculum: we make the task easier for the first few epochs and then gradually make it harder. We warm up the adversarial perturbation budget ($\eps$) by linearly increasing $\eps$ for a fixed number of warm-up epochs. Our results (\cref{tab:eps-warmup}) show that using 20 epochs as warm-up duration gives a significant increase in both clean accuracy and APGD-CE accuracy. This suggests that, while gradually increasing the difficulty of the task does not prove useful for models with larger inductive bias, such as ResNets~\cite{pang2020bag}, it can help for models that have a low inductive bias (e.g., ViTs).

\smallskip \noindent \textbf{Weight decay.}
As pointed out by \citet{pang2020bag}, weight decay has an important role to make models more robust: a larger weight decay helps reduce the generalization gap for robust accuracy. For this reason, we ablate several values of weight decay, in different orders of magnitude. The weight decay used to train XCiT originally was $0.05$~\cite{elnouby2021xcit}, but we get the best results with the weight decay equal to $0.5$ (\cref{tab:weight-decay}), both in terms of clean and robust accuracy. We hypothesize that the regularization introduced by the larger weight decay helps as we have removed heavy data augmentation.

\section{Validating our training recipe at scale}\label{sec:val-at-scale}

We now test the best setup found in the previous section on the full ImageNet-1k dataset, with a range of architectures and model sizes. We show that not only our recipe achieves strong results for XCiT-S when compared to the canonical recipe, but it also enables better results on other modern architectures, and other model sizes in the case of XCiT. Finally, we show that we can also pre-train and fine-tune transformers using this recipe for a larger attack budget: XCiTs pre-trained in this way can be robustly fine-tuned to a diverse set of downstream datasets achieving competitive performance.

\smallskip \noindent \textbf{Our training recipe.} Based on our previous findings, we use a) a 10-epochs, linear warm-up for $\eps$, b) better-tuned data augmentation, by using just weak data augmentation (i.e., random resize, crop and horizontal flipping, and color jitter), and c) 0.5 weight decay. We do a 10 epochs warm-up, instead of 20 (as what \cref{tab:eps-warmup} would suggest), because ImageNet-1k is 10 times larger than ImageNet-100: after 10 epochs the model will have seen enough \textit{easy} images with small perturbations, to then continue the training with the full perturbation budget.

\begin{Summary}[title=\textbf{Proposed training recipe}]{}{}
\begin{itemize}[leftmargin=0pt]
    \item 10-epoch Linear $\eps$-warmup
    \item Only basic data augmentation (random-resize-and-crop + horizontal-flipping + color-jitter).
    \item High weight decay ($0.5$ using AdamW on ImageNet-1k)
\end{itemize}
\end{Summary}

\begin{table}[!htb]
\renewcommand{\arraystretch}{1.2}
\centering
\caption{\textbf{Weak data augmentation with large weight decay is better than heavy data augmentation curriculum.} In adversarial training of ViTs, \textcite{bai2021transformers} previously recommended the use of strong data augmentation, but with a progressive curriculum. We observe that using \textit{only} weak data augmentations strategies, together with larger weight decay, brings better results than using progressive strong data augmentation increased in the first epochs. For a fair comparison, we use identical network architecture and training setup as \textcite{bai2021transformers}. The model marked with $^\dagger$, trained by \textcite{bai2021transformers} with the canonical recipe, failed the training, while our implementation of the same model is marked with $^*$.}
\label{tab:bai-comparison}
    \begin{tabularx}{\linewidth}{*{1}{X}*{2}{c}}
    \toprule
    \multirow{2}{*}{Model}                    & \multicolumn{2}{c}{Accuracy} \\ \cmidrule{2-3}
                                              & Clean       & AutoAttack     \\ \midrule
    GELU ResNet-50~\cite{bai2021transformers} \textit{(c)} & $67.38$     & $35.53$        \\
    DeiT-S$^\dagger$~\cite{bai2021transformers} \textit{(c)}         & ---     & ---        \\
    DeiT-S~\cite{bai2021transformers} \textit{(c + heavy augmentation curriculum)}         & $66.50$     & $35.50$        \\
    DeiT-S$^*$ \textit{(c, ours)}                          & $66.30$     & $32.70$ \\    
    \textbf{DeiT-S} \textit{(ours)}                          & $66.80$     & $37.90$        \\
    \textbf{XCiT-S12} \textit{(ours)}                         & $72.34$     & $41.78$        \\
    \bottomrule
    \end{tabularx}
\end{table}

\subsection{Validating success on the full ImageNet-1k dataset}\label{sec:im-validation}

\noindent \textbf{Intuition of why the recipe should scale up.} We observe that the data augmentation setup, as well as weight decay, could interact with the dataset size: a smaller dataset may need stronger data augmentation, as the model could overfit. In particular, this was experimentally validated for ViTs by \citet{steiner2021train} and \citet{touvron2021training}. On the other side, we note that we use \emph{minimal} data augmentation on the smaller ImageNet-100. Hence, we argue that, as ImageNet-1k is larger than ImageNet-100, it should require at most the same (if not weaker) data augmentation to improve generalization. Regarding weight decay, similarly, for a smaller dataset, we may require a larger weight decay to avoid overfitting, while a smaller weight decay may be desirable for a larger dataset to avoid underfitting. However, as we will show in the next paragraph, using a larger weight decay brings improvements.

\begin{table*}[ht]
    \caption{\textbf{Success on full scale ImageNet-1k dataset.} We validate our method on the full ImageNet-1k dataset in two steps. In table (a) we step-by-step add each finding from our ablation study (Table~\ref{tab:big_ablation}) and shows our bag-of-tricks for ViT's adversarial training generalize to the full ImageNet-1k dataset. The most noticeable improvement comes from using weaker data augmentations, the key finding uncovered in our ablation study. In table (b) we test whether weak data augmentation consistently benefits adversarial training across ViTs or ViTs-inspired architectures. For each network, we consider the training recipe used in the original paper for standard training and compare it with the proposed training recipe. Across all networks, we find that proposed training improves both clean and robust accuracy. We use AutoAttack to measure robust accuracy.}
    \begin{subtable}[t]{\linewidth}
    	\renewcommand{\arraystretch}{1.2}
    	\centering
    	\caption{\textbf{Step-by-step improvements.} The effect of our training recipe components incrementally when tested on the full ImageNet-1k dataset. Overall, our training recipe improves the robust accuracy by $13.08\%$, not at the expense of the clean one, which increases by $0.66\%$.}\label{tab:final-ablation}
        \begin{tabularx}{0.5\linewidth}{*{1}{X}*{2}{c}} \toprule
    		\multirow{2}{*}{Feature}        & \multicolumn{2}{c}{Accuracy}                            \\ \cmidrule{2-3}
    		                                & Clean                      & AutoAttack                 \\ \midrule
    		\textit{XCiT-S12}               & $71.68$                    & $28.70$                    \\ \midrule
    		+ $\eps$ warmup (10 epochs)    & $71.98$ ($+0.30$)          & $29.36$ ($+0.66$)          \\ \midrule
    		+ Tuned data augmentation       & $71.70$ ($-0.28$)          & $38.78$ ($+9.42$)          \\ \midrule
    		+ Tuned weight decay            & $\mathbf{72.34}$ ($+0.64$) & $\mathbf{41.78}$ ($+3.00$) \\ \bottomrule
    	\end{tabularx}
    \end{subtable}
    \newline
    \vspace{0.1pt}
    \newline
    \begin{subtable}[t]{\linewidth}
    \centering
    \caption{\textbf{Cross-architecture generalization.} We compare three different architectures adversarially trained on ImageNet-1k with and without heavy data augmentation and small weight decay. For all three architectures, using weak data augmentation brings an advantage, as opposed to standard training, where heavy data augmentation and smaller weight decay bring an advantage. Since for standard training there is no $\eps$ involved, the \textit{$\eps$ schedule} column applies only to the adversarially trained models. The PoolFormer without data augmentation (marked with $^*$) is trained with $0.05$ weight decay, as with $0.5$ the training collapses. The symbol $^\dagger$ means that the training run collapsed.}\label{tab:recipe-improvement}
    \begin{tabular}{lcccccc}
        \toprule
        \multirow{2}{*}{Architecture}   & \multicolumn{1}{c}{$\eps$ schedule} & \multicolumn{1}{c}{\multirow{2}{*}{\begin{tabular}[c]{@{}c@{}}Heavy data\\ Augmentation\end{tabular}}} & \multicolumn{1}{c}{\multirow{2}{*}{\begin{tabular}[c]{@{}c@{}}Weight\\ Decay\end{tabular}}}   & \multicolumn{2}{c}{Adversarial training} & Standard training  \\ \cmidrule{5-7}
                                        & \multicolumn{1}{c}{}                                                                                   & \multicolumn{1}{c}{}                                                                                   & \multicolumn{1}{c}{}                                                                          & Clean accuracy      & AutoAttack accuracy  & Clean accuracy   \\ \midrule
        \multirow{2}{*}{XCiT-S12}       & \xmark                                                                                                 & \cmark                                                                                                 & $0.05$                                                                                          & $71.68$             & $28.70$              & $\mathbf{80.53}$ \\
                                        & \cmark                                                                                                 & \xmark                                                                                                 & $0.5$                                                                                           & $\mathbf{72.34}$    & $\mathbf{41.78}$     & $78.96$          \\ \midrule
        \multirow{2}{*}{DeiT-S}         & \xmark                                                                                                 & \cmark                                                                                                 & $0.05$                                                                                          & $66.30$             & $32.70$              & $\mathbf{74.61}$ \\
                                        & \cmark                                                                                                 & \xmark                                                                                                 & $0.5$                                                                                           & $\mathbf{66.80}$    & $\mathbf{37.90}$     & $73.38$          \\ \midrule
        \multirow{2}{*}{ConvNeXt-T}     & \xmark                                                                                                 & \cmark                                                                                                 & $0.05$                                                                                          & $0.08^\dagger$      & $0.08^\dagger$       & $\mathbf{79.87}$             \\
                                        & \cmark                                                                                                 & \xmark                                                                                                 & $0.5$                                                                                           & $\mathbf{71.64}$    & $\mathbf{44.44}$     & $77.70$          \\ \midrule
        \multirow{2}{*}{PoolFormer-M12} & \xmark                                                                                                 & \cmark                                                                                                 & $0.05$                                                                                          & $65.88$             & $34.08$              & $\mathbf{76.84}$ \\
                                        & \cmark                                                                                                 & \xmark                                                                                                 & $0.05^*$                                                                                          & $\mathbf{66.16}$    & $\mathbf{34.72}$     & $75.74$          \\ \bottomrule
    \end{tabular}
\end{subtable}
\end{table*}

\smallskip \noindent \textbf{Validating our proposed recipe step-by-step (Table~\ref{tab:final-ablation}).} We find that all three strategies, i.e., using $\eps$ warm-up, weak data augmentation, and large weight decay, helps the performance of adversarial training for the XCiT transformer model. Among them, reducing data augmentation yields the highest benefit, while all aspects combined improve the robust accuracy by $12.42$\%, i.e., by more than $40\%$ relative to the canonical recipe robust accuracy. In comparison, at the time of submission, the leading entry from RobustBench leaderboard for ImageNet-1k~\cite{robustbench}, by \textcite{salman2020adversarially} achieves $38.14\%$ robust accuracy ($3.64$\% lower than ours) and $68.46\%$ clean accuracy ($3.88$\% lower than ours). Note that~\citet{salman2020adversarially} use WideResNet-50-2, which has $2.6\times$ more parameters and $2.4\times$ more FLOPs than XCiT-S12. Finally, we validate our choice not to use RandomErasing to keep the recipe light: the model trained with it only has $40.60\%$ robust accuracy ($1.18\%$ less than without), and comparable clean accuracy ($72.42\%$).

\begin{figure}
    \centering
    \includegraphics[width=0.7\linewidth]{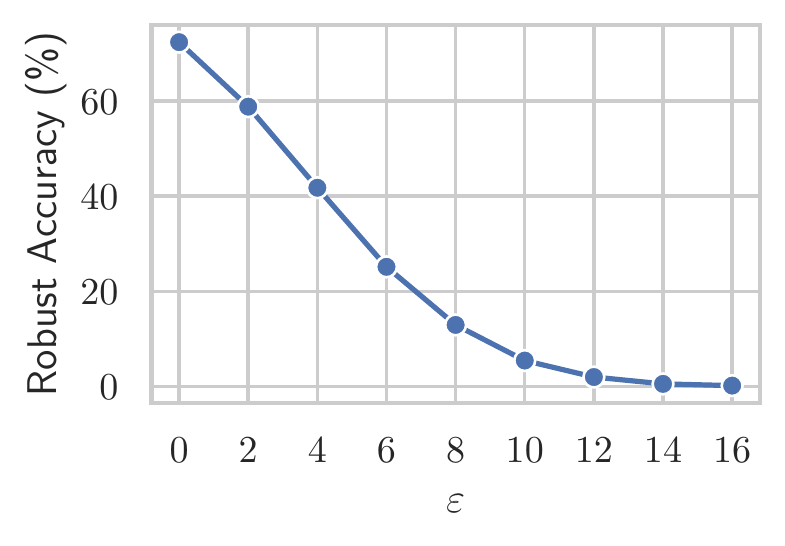}
    \caption{\textbf{AutoAttack is reliable.} We show how the robustness of XCiT-S trained for $\eps=\nicefrac{4}{255}$ goes down as $\eps$ increases. As expected from a reliable attack, the robust accuracy monotonically, but gently decreases with $\eps$.}
    \label{fig:aa-check}
\end{figure}

\smallskip \noindent \textbf{AutoAttack reliability.} To make sure that AutoAttack is a reliable attack for XCiT as much as it is for adversarially trained ResNets (i.e., the model does not suffer from gradient masking or other factors which may impact AutoAttack's performance), we study how the robustness of XCiT-S12 changes when $\eps$ increases. As we can see from \cref{fig:aa-check}, the robust accuracy decreases monotonically -- but slowly -- until it reaches $\sim 0\%$ when $\eps = 16$. This suggests that AutoAttack has no clear issues to find adversarial examples for XCiT. Finally, we note that, among the $5000$ validation images used by RobustBench, $2345$ are not perturbed successfully by either of APGD-CE and APGD-T~\cite{croce2020reliable}, and among those $2345$, only $1$ is perturbed successfully by the FAB-T~\cite{croce2020minimally} attack, and among the $2344$ remaining images, none is successfully perturbed by the black-box Square attack~\cite{andriushchenko2020square}. The fact that the black-box attack does not manage to decrease the robust accuracy of the white-box ones further suggests that no issues are encountered when computing gradients with white-box attacks. For completeness, we also report PGD-\{$5$, $10$, $50$, $100$\} results in \cref{tab:pgd-results} in the appendix.

\smallskip \noindent \textbf{Comparison with Bai \textit{et al.}~\cite{bai2021transformers}.} In their work, \textcite{bai2021transformers} observe that training DeiT-S with strong augmentations leads to a collapse of the training procedure, and training with no strong augmentation at all (but with no changes in terms of weight decay) leads to suboptimal performance. On the other hand, they find that increasing the intensity of data augmentation in the first ten epochs stabilizes the training procedure and leads to $35.50\%$ AutoAttack accuracy. However, we observe that, using our implementation, we can train a DeiT-S with heavy augmentation to nontrivial AutoAttack accuracy ($32.70\%$). Our hypothesis for this difference is that we used a 10-epochs learning rate warm-up, instead of a 5-epochs one. Moreover, using our recipe, DeiT-S achieves better robust accuracy than when using the canonical recipe with an increase in data augmentation intensity. Finally, we improve over the best result reported in their work, obtained with a GELU ResNet-50 with both better clean and robust accuracy. The comparison is summarized in \cref{tab:bai-comparison}.

\smallskip \noindent \textbf{Our recipe's benefits generalize across architectures (Table~\ref{tab:recipe-improvement}).} In adversarial training, reducing data augmentation strength yields benefits not only for transformer architectures, such as XCiT and PoolFormer~\citep{yu2021metaformer} but also for the ConvNeXt~\citep{liu2022convnet} CNN architecture, as we summarize in \cref{tab:recipe-improvement}. With ConvNeXt-T, when using heavy data augmentation as in the original paper, both with, and without, $\eps$ warm-up, the training fails. However, when trained using our recipe, i.e., with larger weight decay and weak data augmentation, the model achieves state-of-the-art results. This phenomenon is likely due to the fact that adversarial training requires much higher capacity networks than standard training~\cite{madry2017towards}, as it solves the learning objective on all samples within the perturbation budget. While solving it on heavily augmented instances, the network sacrifices network capacity but doesn't yield an equivalent benefit in generalization. On the other hand, from the PoolFormer family, we train PoolFormer-M12, which is both smaller and has fewer FLOPs than ResNet-50, and we observe that, if we use the weight decay from our recipe (0.5) it is extremely unstable. Hence, we use our training recipe with the canonical weight decay (0.05). If we train the same model with full data augmentation from the canonical recipe (the same as the DeiT and XCiT one), we obtain a model with both worse AutoAttack and clean accuracy. Our recipe then brings an improvement also for this architecture, albeit by a smaller margin than ConvNeXt-T.

\begin{table}[h]
\footnotesize
\renewcommand{\arraystretch}{1.2}
\centering
\caption{\textbf{Performance in the $\ell_2$ threat model.} We train XCiT-S12 to be robust against $\ell_2$-bounded perturbation with $\eps=3.0$ employing the canonical recipe and compare it with our recipe. We show that despite the XCiT-S12 trained with our recipe having slightly lower clean accuracy, it has significantly larger robust accuracy. We also show the accuracies of both a GELU ResNet-50 trained according to the set-up of \textcite{bai2021transformers} and the ReLU ResNet-50 shared by \textcite{salman2020adversarially}. Both have lower performance than the XCiT-S12. The symbol $^\dagger$ means that the training run collapsed.}\label{tab:l2-robustness}
\begin{tabular}{lccc}
\toprule
\multirow{2}{*}{Model}    & \multirow{2}{*}{Recipe}     & \multicolumn{2}{c}{Accuracy} \\ \cmidrule(l){3-4} 
                          &                             & Clean       & AutoAttack     \\ \midrule
ReLU ResNet-50~\cite{salman2020adversarially}           & \textit{Canonical}      & $62.86$     & $34.84$        \\
GELU ResNet-50             & \textit{Canonical}         & $66.14$       & $35.60$        \\ \midrule
\multirow{2}{*}{XCiT-S12} & \textit{Canonical}          & $\mathbf{71.24}$       & $29.38$        \\
                          & \textit{Ours}               & $70.78$       & $\mathbf{39.94}$        \\ \midrule
\multirow{2}{*}{ConvNeXt-T} & \textit{Canonical}$^\dagger$        & ---           & ---            \\
                            & \textit{Ours}             & $\mathbf{70.58}$       & $\mathbf{41.44}$        \\ \midrule
\multirow{2}{*}{PoolFormer-M12} & \textit{Canonical}    & $65.26$       & $32.10$        \\
                                & \textit{Ours}         & $\mathbf{66.40}$       & $\mathbf{36.04}$        \\ \midrule
\multirow{2}{*}{DeiT-S}   & \textit{Canonical}          & $64.20$       & $31.10$        \\
                          & \textit{Ours}               & $\mathbf{66.64}$       & $\mathbf{36.20}$        \\ \bottomrule
\end{tabular}
\end{table}

\smallskip \noindent \textbf{Our recipe's benefits generalize to the $\ell_2$ threat model (\cref{tab:l2-robustness}).} Even though the $\ell_\infty$ threat model is the scenario that has been studied the most so far for ImageNet~\cite{croce2020robustbench,bai2021transformers,mao2022easyrobust,wong2020fast}, it is also important to see whether our recipe generalizes to other threat models, as they may bring additional benefits, such as better standard fine-tuning performance~\cite{salman2020adversarially}. For this reason, we train XCiT-S12 with both the canonical and our recipe to be robust against $\ell_2$-bounded perturbation with $\eps=3.0$. While the model trained with our recipe has slightly lower clean accuracy, the robust accuracy is better by one-third (i.e., $29.38\%$ vs. $39.94\%$)., bringing a significant improvement. This suggests that our recipe also generalizes to the $\ell_2$ threat model for XCiT-S12. For completeness, we also train a GELU ResNet-50 using the recipe of \textcite{bai2021transformers} and report the results of the ReLU ResNet-50 shared by \textcite{salman2020adversarially}. Our XCiT-S12 performs better than both, by a substantial margin. We can also observe that our recipe brings improvements across the board for other architectures: ConvNeXt-T, PoolFormer-M12, and DeiT-S.

\begin{table}[t]
    \footnotesize
    \renewcommand{\arraystretch}{1.2}
    \centering
    \caption{\textbf{Scaling to larger models.} We test our recipe on larger variants XCiT and compare it to robust ResNets. The XCiT variants outperform ResNets by a wide margin, and achieve \textit{top} rank on the RobustBench~\citep{croce2020robustbench} benchmark. Finally, we use our training recipe for ConvNeXT~\cite{liu2022convnet}, PoolFormer~\cite{yu2021metaformer}, and DeiT~\cite{touvron2021training}. We use baseline results from \citet{bai2021transformers,salman2020adversarially}.}\label{tab:scaling-up}
    \begin{tabular}{lcccc} \toprule
    	\multirow{2}{*}{Architecture} & \multirow{2}{*}{Parameters} & \multirow{2}{*}{GFLOPs} & \multicolumn{2}{c}{Accuracy} \\ \cmidrule{4-5}
                                                        &        &          & Clean     & AutoAttack  \\ \midrule
    	GELU ResNet-50~\cite{bai2021transformers}       & $25$M  & $4.11$   & $67.38$   & $35.51$     \\
        WideResNet-50-2~\cite{salman2020adversarially}  & $68$M  & $11.47$  & $68.46$   & $38.14$     \\
        \midrule
    	XCiT-S12                                        & $26$M  & $4.82$   & $72.34$   & $41.78$     \\
    	XCiT-M12                                        & $46$M  & $8.54$   & $74.04$   & $45.24$     \\
    	XCiT-L12                                        & $104$M & $18.97$  & $73.76$   & $47.60$     \\ \midrule
        PoolFormer-M12                                  & $22$M  & $3.22$   & $66.16$   & $34.72$     \\
        DeiT-S                                          & $22$M  & $4.61$   & $66.80$   & $37.90$     \\
        ConvNeXt-T                                      & $29$M  & $4.50$   & $71.64$   & $44.44$     \\ \bottomrule
    \end{tabular}
\end{table}

\begin{figure}[t]
    \centering
    \begin{subfigure}[t]{0.46\linewidth}
        \centering
        \includegraphics[width=\linewidth]{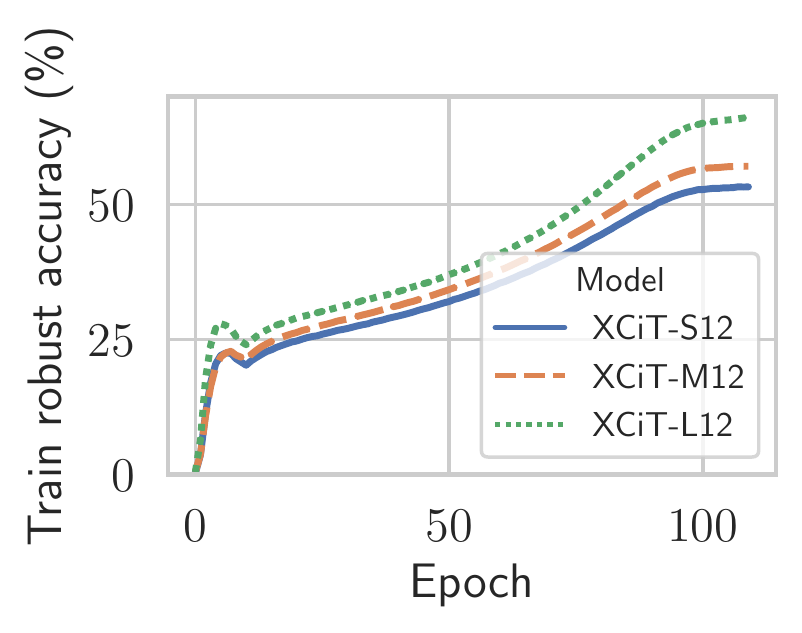}
    \end{subfigure}
    \hspace{0.02\textwidth}
    \begin{subfigure}[t]{0.46\linewidth}
        \centering
        \includegraphics[width=\linewidth]{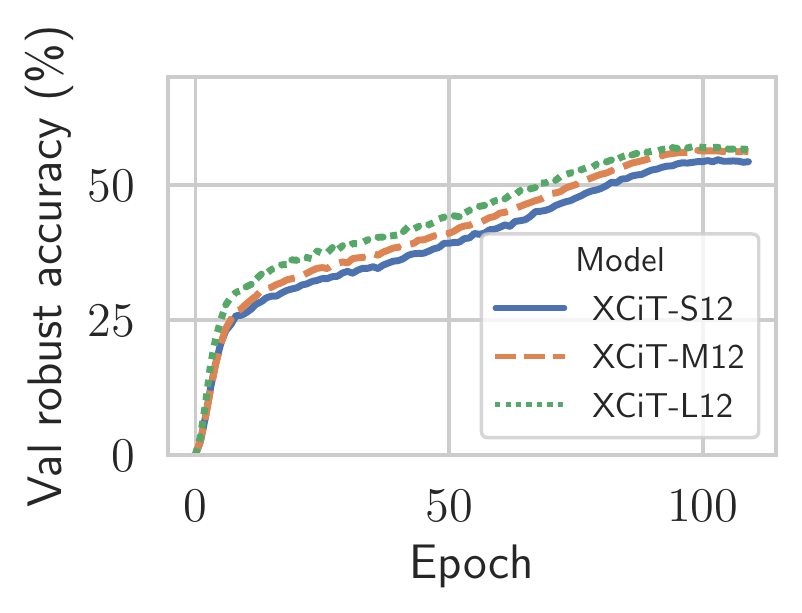}
    \end{subfigure}
    \caption{\textbf{Learning curves for the XCiT models.} We show the progress of the training \figleft{} and validation \figright{} FGSM accuracy for XCiT-\{S,M,L\}12 trained with our recipe.}\label{fig:learning-curves}
\end{figure}

\smallskip \noindent \textbf{Validating success on large-scale models (\cref{tab:scaling-up}).} Given that our training recipe successfully generalizes across network architectures, we use it to train larger-scale models. For this experiment, we train on a 64-core TPUv4 pod, while scaling batch size and learning rate accordingly. The total training time for XCiT-S12 is 19h30m, for XCiT-M12 it is 33h, and for XCiT-L12 it is 39h. We provide our detailed setup in \cref{sec:setup}, and we show the progress of the validation FGSM accuracy in \cref{fig:learning-curves}. When increasing the scale of the model from S12 to M12, we observe consistent improvement in robust accuracy (from $41.78$\% to $45.24$\%). Regarding XCiT-L12, instead, we first observe that, when trained with 1-step FGSM, it achieves sub-par robust accuracy ($43.78$\%). Investigating more, we note that XCiT-L12 has better PGD-10 accuracy than XCiT-M12 ($52.22$\% vs.\ $51.50$\%) on the RobustBench ImageNet-1k subset. However, when it comes to APGD-CE (the first of the AutoAttack ensemble), XCiT-L12 is outperformed by XCiT-M12 ($46.14$\% vs.\ $47.58$\%). For this reason, we train XCiT-L12 using 2-step FGSM, which results in a model with better robust accuracy than XCiT-M12 ($47.60$\% vs. $45.24$\%). Finally, to understand the accuracy-robustness trade-off, we also standardly train these networks for 100 epochs. XCiT-S12, M12, and L12 achieve $80.36$\%, $81.71$\%, and $82.65$\% clean accuracy, respectively. This suggests that adversarial training in ViT sacrifices $\sim 10$\% clean accuracy to achieve robustness.

\smallskip \noindent \textbf{Comparison with the EasyRobust library models.} \textcite{mao2022easyrobust} recently released EasyRobust, a PyTorch library to perform adversarial training. Concurrently to our work, they released several models on GitHub (albeit without a paper that gives information about how they trained the models). These models have been trained with a variety of recipes\footnote{We discussed the parameters with the authors via e-mail.}. In particular, ViT-S has been trained with the \textit{canonical} recipe, while the other models in \cref{tab:scaling-up} have been trained with the same data augmentation we use in our recipe, plus the \emph{Lightning noise} data augmentation, used in the \texttt{robustness} library~\cite{robustness}. All the models have been trained for 300 epochs. As they have been trained for $3\times$ the number of epochs as ours, it is not straightforward to draw a comparison. In any case, we note that their top-performing models (when taking into account FLOPs and parameters), i.e., Swin-S ($73.41\%$ clean and $46.76\%$ AutoAttack accuracy) and Swin-B ($75.05\%$ clean and $47.42\%$ AutoAttack accuracy), have been trained without strong data augmentations and show very promising results. We leave for future research on whether integrating the other elements of our recipe (i.e., $\eps$ warm-up and larger weight decay) can further improve the performance of these models.

\subsection{Beyond pre-training: Success of our approach with transfer learning}

\smallskip \noindent \textbf{Setup.}
We consider four datasets, namely CIFAR-10 and CIFAR-100~\cite{krizhevsky09learning}, Caltech-101~\cite{feifei2004learning}, and Oxford flowers~\cite{nilsback08automated}. These datasets cover a diverse range in terms of the number of images and image resolution. For all datasets, we use $\eps = \nicefrac{8}{255}$ in both pre-training and finetuning. Our baselines will be XCiT-S12 and ResNet-50 pre-trained with the canonical recipe, which we compare their success with networks pre-trained with our proposed recipe. Additional details about pre-training, as well as the performance of the resulting models, can be found in \cref{sec:setup}. When finetuning, we use the identical recipe as pre-training for the high-resolution dataset, and a slightly different one, employing TRADES-10~\cite{zhang2019theoretically}, for the low-resolution ones. We provide extensive details on our finetuning procedure in \cref{sec:setup}.

\begin{table*}[h]
    \caption{\textbf{Advantages of the pre-training recipe also directly transfer to fine-tuning.} When finetuning results on both high and low-resolution datasets, we find that our proposed recipe achieves better performance. The XCiT-S12 marked with \textit{(c)} is the one pre-trained using the canonical recipe used for standard training, while \textit{(ours)} refers to models pre-trained with our proposed recipe. Both models are adapted by changing the stride of the initial convolution to change the patch size. Note that pre-training and finetuning methods (standard/adversarial) are kept identical for Caltech-101 and Oxford Flowers. The ResNet-50 pre-trained checkpoint is from \textcite{salman2020adversarially}, and is adapted to small resolutions by changing the stride of the first convolution layer, and the WideResNet-28-10 in \cref{tab:cifar-10-ft,tab:cifar-100-ft} are from \citet{hendrycks19using}.}
    \begin{subtable}[t]{0.52\linewidth}
    	\centering
    	\footnotesize
    	\renewcommand{\arraystretch}{1.2}
    	\caption{\textbf{Fine-tuning on high-resolution datasets.} }\label{tab:hi-res-finetuning}
    	\resizebox{\linewidth}{!}{
    	\begin{tabular}{cccccc}
    		\toprule
    		\multirow{3}{*}{Fine-tuning} & \multirow{3}{*}{Model} & \multicolumn{4}{c}{Dataset}                                                       \\ \cmidrule{3-6}
    		                             &                        & \multicolumn{2}{c}{Caltech-101}             & \multicolumn{2}{c}{Oxford Flowers}  \\ \cmidrule{3-6}
    		                             &                        & Clean                  & AutoAttack         & Clean            & AutoAttack       \\ \midrule
            \multirow{3}{*}{Standard}    & ResNet-50              & $86.97$                & $7.56$             & $86.28$          & $1.96$           \\
                                         & XCiT-S12 \textit{(c)}         & $89.92$                & $0.96$             & $88.13$          & $0.99$           \\
    		                             & XCiT-S12 \textit{(ours)}         & $\mathbf{90.36}$       & $\mathbf{17.12}$   & $\mathbf{91.65}$ & $\mathbf{5.71}$  \\ \midrule
    		\multirow{3}{*}{Adversarial} & ResNet-50              & $81.38$                & $34.49$            & $74.51$          & $32.75$          \\
    		                             & XCiT-S12 \textit{(c)}         & $86.18$                & $58.84$            & $76.26$          & $42.42$          \\
    		                             & XCiT-S12 \textit{(ours)}         & $\mathbf{87.59}$       & $\mathbf{61.74}$   & $\mathbf{82.86}$ & $\mathbf{47.91}$ \\ \bottomrule
    	\end{tabular}}
    	\vspace{5pt}
	\end{subtable}
	\hfill
    \begin{subtable}[t]{0.45\linewidth}
        \centering
        \caption{\textbf{CIFAR-10 adversarial fine-tuning.}}\label{tab:cifar-10-ft}
        \renewcommand{\arraystretch}{1.3}
        \begin{tabular}{lcc} \toprule
    		Model               & Clean Accuracy & AA Accuracy \\ \midrule
    		WideResNet-28-10~\cite{hendrycks19using}    & $87.11$          & $54.92$       \\
    		ResNet-50           & $84.80$          & $41.56$       \\
    		XCiT-S12 \textit{(c)}      & $89.07$          & $54.37$       \\
    		XCiT-S12 \textit{(ours)}      & $90.06$          & $56.14$       \\
    		XCiT-M12 \textit{(ours)}      & $91.30$          & $57.27$       \\
    		XCiT-L12 \textit{(ours)}      & $\textbf{91.73}$          & $\textbf{57.58}$       \\
    		\bottomrule
    	\end{tabular}
    \end{subtable}
    \hfill
    \begin{subtable}[t]{0.52\linewidth}
    	\centering
    	\footnotesize
    	\caption{\textbf{Pre-training is necessary for smaller datasets.} Fine-tuning performance when a XCiT-S12 model is 1) trained from scratch 2) trained from scratch with extra synthetic data~\cite{sehwag2021robust} 3) pre-trained on ImageNet-1k.}\label{tab:cifar-10-finetuning}
    	\renewcommand{\arraystretch}{1.2}
    	\begin{tabular}{cccc}
    		\toprule
    		\multirow{2}{*}{Pre-training}       & \multirow{2}{*}{Synthetic data}       & \multicolumn{2}{c}{Accuracy}          \\
    		\cmidrule{3-4}
    		                                    &                                       & Clean Accuracy      & AA Accuracy     \\
    		\midrule
    		\xmark                              & \xmark                                & $82.84$             & $39.49$         \\
    		\xmark                              & \cmark                                & $80.01$             & $47.88$         \\
    		\cmark                              & \xmark                                & $\mathbf{90.06}$    & $\mathbf{56.14}$\\
    		\bottomrule
    	\end{tabular}
    \end{subtable}
    \hfill
	\begin{subtable}[t]{0.45\linewidth}
	\centering
	    \caption{\textbf{CIFAR-100 adversarial fine-tuning.}}\label{tab:cifar-100-ft}
	    \renewcommand{\arraystretch}{1.34}
	    \begin{tabular}{lcc} \toprule
    		Model                       & Clean Accuracy & AA Accuracy  \\ \midrule
    		WideResNet-28-10~\cite{hendrycks19using}  & $59.23$          & $28.42$    \\
    		ResNet-50                             & $61.28$          & $22.01$    \\
    		XCiT-S12 \textit{(c)}                 & $65.44$          & $30.97$    \\
    		XCiT-S12 \textit{(ours)}              & $67.34$          & $32.19$    \\
    		XCiT-M12 \textit{(ours)}              & $69.21$          & $34.21$    \\
    		XCiT-L12 \textit{(ours)}              & $\textbf{70.76}$          & $\textbf{35.08}$    \\
    		\bottomrule
    	\end{tabular}
	\end{subtable}
\end{table*}

\smallskip \noindent \textbf{Success in finetuning on high resolution datasets.} We finetune our pre-trained networks on both Caltech-101 and Oxford Flowers for 20 epochs. From \cref{tab:hi-res-finetuning} we can see that: 1) we can easily fine-tune on both datasets out-of-the-box. 2) Our XCiT achieves the best results, by a significant margin, for both the standardly and adversarially trained models. 3) We can observe that, for XCiT-S12, there is a very good robustness-accuracy trade-off, as the robust model trained on Caltech-101 has a small drop of $2.77\%$ w.r.t.\ the standardly fine-tuned model (vs.\ a $5.59$\% drop for ResNet-50), and the robust model trained on Oxford Flowers has a drop in terms of clean accuracy of $8.79\%$ (vs.\ an $11.77\%$ drop for ResNet-50). 4) Despite the larger clean accuracy, XCiT-S12 is more robust, having a robust accuracy better by $27.25\%$ on Caltech-101, and $15.16\%$ better on Oxford Flowers. 5) The models pre-trained with our recipe lead to better results across the board when compared to the models pre-trained with the canonical recipe.

\smallskip \noindent \textbf{Adapting to small resolution images.} As we pre-train on the ImageNet-1k dataset at $224\times224$ resolution, we first need to modify the network architecture to adapt to small resolutions, i.e., $32\times32$ for CIFAR-10 and CIFAR-100. Previous work~\citep{shao2021adversarial} achieves this by down-sampling the weights of the convolutional layer that is used by ViT to embed each patch. For XCiT, we adapt the patch-processing module, which converts patches to 1-D vectors, by changing the stride of subsequent convolutions from two to one. We finetune the networks for 20 epochs and report our results in \cref{tab:cifar-10-ft,tab:cifar-100-ft}. We provide additional details about the models' adaptation and fine-tuning hyper-parameters in \cref{sec:setup}. To highlight the necessity of pre-training, we also train from scratch on CIFAR-10 using the same setup without pre-training for 300 epochs. We also do a training run using additional synthetic data from \citet{sehwag2021robust} to compensate for the smaller size of the dataset.
Moreover, we compare our models to the WideResNet-28-10 by~\citet{hendrycks19using}, who fine-tune a model pre-trained on a sub-sampled version of ImageNet-1k, and to the fine-tuned ResNet-50 pre-trained by~\citet{salman2020adversarially}. Our models perform better than both baselines. In particular, we suspect the performance of the ResNet-50 to be sub-optimal because of the mismatch in resolution, while this factor may not affect XCiT as much. As a matter of fact, XCiT is shown to be more resilient to changes in image resolution~\cite{elnouby2021xcit}, while \citet{salman2020adversarially}, when using their pre-trained model for standardly fine-tuning, up-sample both CIFAR datasets. 
Finally, we note that our XCiT-L model fine-tuned on CIFAR-100, at the time of the submission, would rank second in the corresponding RobustBench leaderboard~\cite{croce2020robustbench}, with better performance than more compute-intensive architectures.

\section{Understanding the success of Vision Transformers in adversarial training}\label{sec:understanding}

So far, we have shown that changing architecture can improve adversarial robustness by a large margin when using an appropriate training recipe. Now, we first investigate a potential reason for this: we hypothesize that, for the top-performing models, few-step attacks are more effective than a conventional ResNet or an XCiT trained with the canonical recipe. This \emph{attack effectiveness} makes the models more robust at test time, as the models have seen stronger attacks during training. After that, we further explore one consequence of robust models: we propose a way to \emph{quantify} the semantic nature of adversarial perturbations. We show that the perturbations targeting a robust XCiT-S have more semantic features than those targeting a robust GELU ResNet-50, reflecting the fact that the robust XCiT-S has larger robust accuracy than the robust ResNet-50.

\subsection{Attack effectiveness influences adversarial training}\label{sec:attack-effectiveness}
Adversarial training (\cref{eq:adv-train}) solves a $\min$-$\max$ optimization where the inner maximization aims to generate strong adversarial examples, while the outer minimization optimizes the network parameters to correctly classify these adversarial examples. The choice of the network architecture simultaneously impacts the optimization success in both $\min$ and $\max$ problems. Better network architectures can certainly achieve better solutions for outer $\min$ problems (e.g., scaling in neural network size leads to better performance). However, their success in adversarial training can simultaneously stem from the ability to achieve a better solution for the inner $\max$ problem, i.e., generate stronger adversarial examples. It is common to use only a few gradient steps~\cite{madry2017towards}, even one in some cases~\cite{wong2020fast}, to reduce the computational burden of adversarial training. Thus the key question is not whether it's \textit{easier} to generate strong adversarial examples for ViTs but whether it's \textit{easier} under few-steps gradient attacks. Our approach in this direction is not without precedent, as previous work on CNNs observed that better robustness with improved architectural components is because they may make it easier to generate effective adversarial examples in a few steps~\cite{xie2020smooth}.

\begin{figure}[!htb]
    \begin{minipage}{.48\linewidth}
        \centering
        \includegraphics[width=\linewidth]{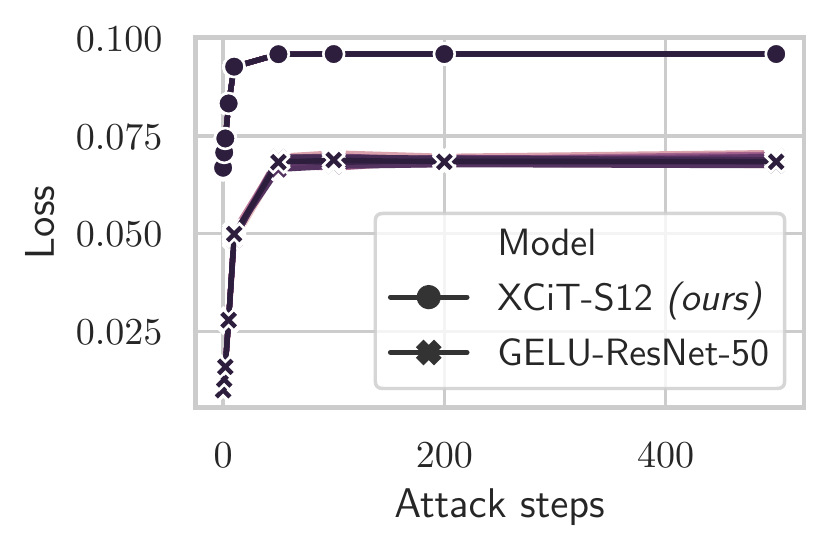}
        \caption{\textbf{Is PGD-200 a good Oracle?} Saturating of the cross-entropy loss in separate runs of PGD attacks with different numbers of steps, perturbing the same input. Plots for additional points are in~\cref{sec:additional-sanity-checks}.}\label{fig:attack-convergence}
    \end{minipage}
    \hfill
    \begin{minipage}{.48\linewidth}
        \centering
        \includegraphics[width=\linewidth]{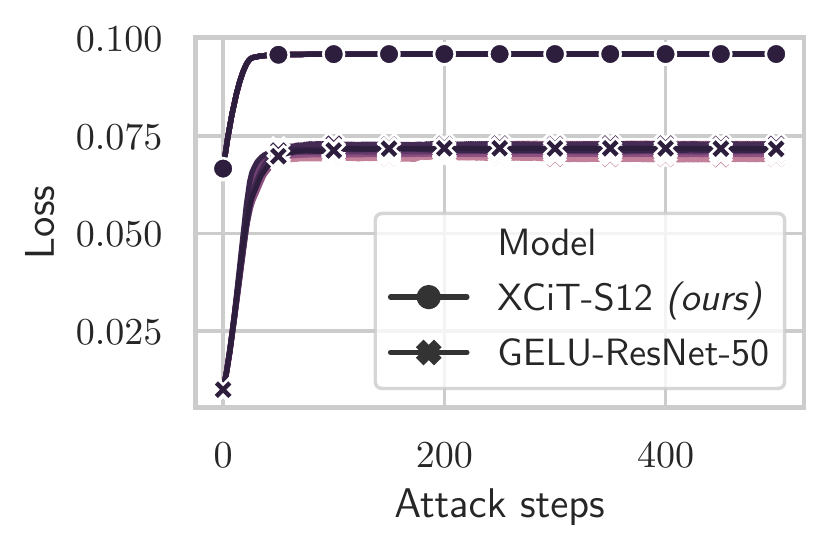}
        \caption{\textbf{Attacks with different initialization converge to very similar losses.} Evolution of the loss for different runs of PGD attacks perturbing the same input, using a large step size. Plots for additional points are in~\cref{sec:additional-sanity-checks}.}\label{fig:attack-convergence-one-run}
    \end{minipage}
\end{figure}

\smallskip \noindent \textbf{Attacking XCiT with gradient-based attacks is as tractable as attacking ResNet.}
While it is empirically well known that it is moderately tractable to optimize adversarial examples with gradient-based methods on adversarially trained ResNets~\cite{madry2017towards}, to the best of our knowledge this has not been studied in the case of adversarially trained ViTs. For this reason, we compute the loss given by separate PGD attacks ran with a different number of steps ($1$, $5$, $10$, $50$, $100$, $200$, $500$), scaling the attack step size accordingly: the maximum loss is reached by attacks that use at least 100 steps. We run this experiment with twenty different random restarts for each point to see if they all converge to similar maxima: all the runs for each point show extremely similar loss curves, for both the robust XCiT-S12 and the robust ResNet-50 (\cref{fig:attack-convergence}). Finally, in \cref{fig:attack-convergence-one-run} we show how the loss changes at every step during twenty separate PGD-500 runs, each starting from a different random start, targeting the same point, with a relatively large step size. This shows that the attack converges after a few steps. We show more results for both experiments on thirty-one more random samples in \cref{sec:additional-sanity-checks}.

\smallskip \noindent \textbf{Effectiveness of $k$-step attacks.} We use the following metric ($d_k$) to measure the tractability of the inner maximization problem with a $k$-step gradient attack, which we call \textit{attack effectiveness}:
\begin{equation}
    d_k = \frac{\Ls(\rvx + \vdelta_{k}, \ry; \vtheta) - \Ls(\rvx + \vdelta_{O}, \ry; \vtheta)}{\Ls(\rvx + \vdelta_{O}, \ry; \vtheta)}
\end{equation}
where $\vdelta_{k}$ is the perturbation generated with $k$ step attacks while $\vdelta_{O}$ is the perturbation generated with an \textit{Oracle}, i.e., a strong attack. As commonly done, we use cross-entropy loss ($\Ls$). This metric measures the strength of a $k$-step attack compared to the strongest attack. As validated previously, using a very high number of attack iterations appears to find the local optimum. Thus we use $200$ attack steps as an Oracle. We measure the effectiveness of adversarial examples generated with PGD-\{$1$, $2$, $5$, $10$\} attacks. We use the full ImageNet-1k validation dataset (50,000 images) for this experiment.

\begin{figure}[!htb]
    \centering
    \begin{subfigure}[t]{0.45\linewidth}
        \centering
        \includegraphics[width=\textwidth]{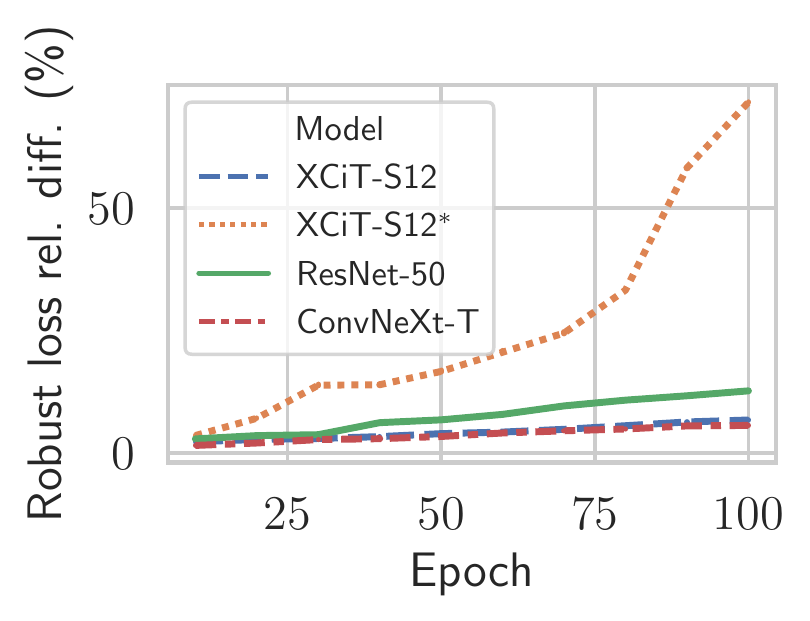}
        \caption{Relative difference between the adversarial loss computed with PGD-1 and the one computed with PGD-200, varying over the training epochs.}
        \label{fig:attack_effectiveness-last_epoch}
    \end{subfigure}
    \hspace{0.02\textwidth}
    \begin{subfigure}[t]{0.45\linewidth}
        \centering
        \includegraphics[width=\textwidth]{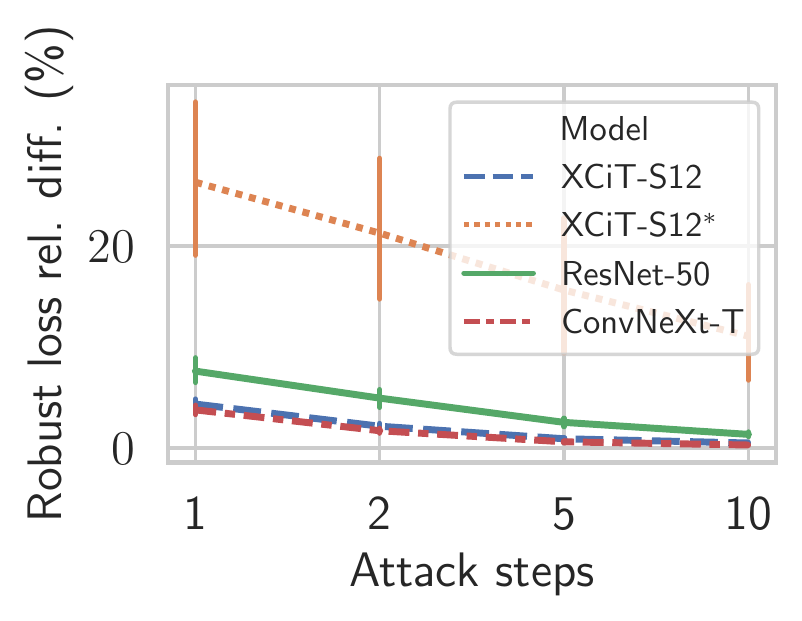}
        \caption{Relative difference between the adversarial losses computed with PGD attacks with 1, 2, 5, and 10 steps, and the adversarial loss computed with PGD-200, averaged across 10 epochs.}
        \label{fig:attack_effectiveness-steps}
    \end{subfigure}
    \caption{\textbf{Few-step attacks of the more robust models are more effective throughout the training}. We test whether the higher robustness of models relates to the ease of optimization of adversarial loss, i.e., using few-step attacks during adversarial training. We measure the relative difference in the success of a few-step attack, i.e., weak but fast, compared to a strong one. We can observe that the relative difference is smaller for the more robust models. This suggests that these few-step attacks (with often one or two steps) are more effective for XCiT-S and ConvNeXt, hence making the final models more robust. We can also observe a significant difference between the XCiT-S12 trained with our recipe and the one trained with the canonical recipe (marked as ``XCiT-S12$^*$'').} These results are computed on the ImageNet-1k validation set and the confidence intervals are over three runs.\label{fig:attack_effectivness}
\end{figure}

\smallskip \noindent \textbf{Few step attacks are highly effective for the most successful models.} Throughout training progress (epochs $1$ to $100$), we find that a single-step attack is more effective against our trained XCiT than a ResNet-50 network (\cref{fig:attack_effectiveness-last_epoch}). A similar trend is observed when we ablate across the number of attack steps (\cref{fig:attack_effectiveness-steps}). This does not hold just for transformers: we observe that the modern ConvNeXt model, which is as successful as XCiT, also enjoys high effectiveness of few-steps attacks. Both observations suggest that the ease of optimizing the inner $\max$ problem with few-step attacks does indeed heavily impact the success of a model in adversarial training.

\smallskip \noindent \textbf{Attack effectiveness in proposed vs canonical training recipe.} Since the $\min$ and $\max$ problems are solved alternatively, i.e., network parameters are continuously updated, the ability the generate strong adversarial examples under fixed steps attacks would vary over time. To analyze this phenomenon, we compare the effectiveness of attacks when attacking the XCiT-S12 trained with the canonical training recipe (with just the $\eps$ warm-up from our recipe): we observe that when using the canonical recipe, the attacks are extremely ineffective, also compared to ResNet's training. This shows that the way we perform the (outer) $\min$ optimization of adversarial training influences the ease of (inner) $\max$ optimization, and an effective training recipe is crucial for successful training.

\smallskip \noindent \textbf{Discussion.} Intuitively, one could argue that it's easy to generate attacks for poorly trained models, and these models show no robust accuracy. However, in our case, the XCiT-S12 trained with our recipe, not only has nontrivial performance, but the clean accuracy is better than that of the XCiT-S12 trained with the canonical recipe. Nonetheless, few-steps attacks generate more powerful adversarial examples for these models, and doing more steps does not bring a significant advantage. On the other hand, for other models, such as the XCiT-S12 trained with the canonical recipe or the GELU ResNet-50, doing more attack steps brings a larger advantage, suggesting that for these models the inner $\max$ optimization is harder.

\subsection{Semantic nature of XCiT's adversarial perturbations}

Earlier works have demonstrated that adversarial perturbations for robust CNNs have semantic, interpretable patterns rather than unintelligible noise as in the case of non-robust networks~\cite{engstrom2019adversarial}. Given the different nature of the ViTs architectures such as XCiT, compared to CNNs, it is natural to ask whether a similar characteristic also emerges for these architectures. Hence, we explore the perturbations targeting a robust XCiT and compare them with those targeting a non-robust XCiT from \texttt{timm}~\cite{rw2019timm}, and those targeted to the robust GELU ResNet-50 by~\citet{bai2021transformers}.

\begin{figure*}[ht]
	\centering
	\includegraphics[width=\linewidth]{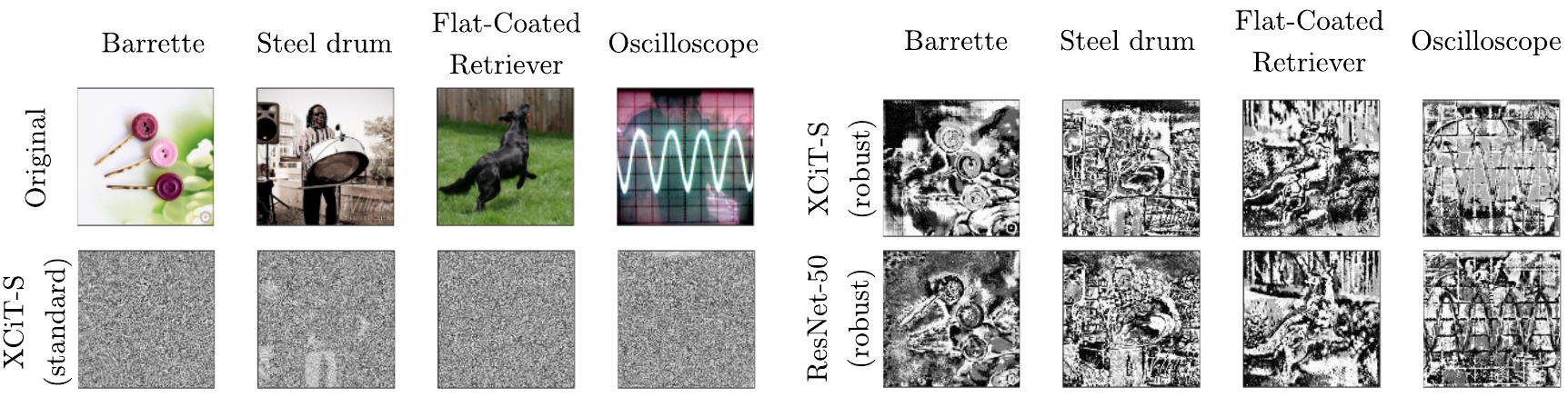}
	\caption{\textbf{Example adversarial perturbations for robust vs. non-robust models.} Comparison between the adversarial perturbations generated for a robust XCiT-S12, a robust GELU ResNet-50 by~\textcite{bai2021transformers} and a regular XCiT-S12 from the timm~\cite{rw2019timm} library. The perturbations, generated with a $\ell_\infty$ attack ($\eps$=$\frac{4}{255}$), are scaled to the $[0, 1]$ range, and, for the sake of visualization, transformed into black and white images. Similar to robust CNNs~\cite{engstrom2019adversarial}, as demonstrated for a ResNet-50 here, robust XCiT perturbations also exhibit semantic characteristics in its adversarial perturbations. When quantitatively compared, we find that XCiT adversarial perturbations have even more semantic information than the robust ResNet-50 perturbation.}\label{fig:attacks}
\end{figure*}

\smallskip \noindent \textbf{Quantifying the semantic nature of perturbations.}
To quantify how semantic the perturbations are, we propose to classify untargeted adversarial perturbations with high-performing, standardly trained models. We hypothesize that if a perturbation is semantic enough (i.e., tries to change the nature of the input from the point of view of the human eye), then it should be classified with the class it tries to evade, as the perturbations should be focused on the shapes characterizing such class in the input image. We use, as classifiers, ConvNeXt-XL~\citep{liu2022convnet}, with $87.01\%$ clean accuracy, BeiT-L~\citep{bao2022beit}, with $87.48\%$ clean accuracy, and Swin-B~\citep{liu2021swin}, with $86.32\%$ clean accuracy on ImageNet-1k, using the implementation and pre-trained weights from the timm library~\citep{rw2019timm}. All the models accept input size 224$\times$224. We generate PGD-100 perturbations for the 5000 images subset of RobustBench for our robust XCiT-S12 and a non-robust XCiT-S12 with pre-trained weights from timm~\citep{rw2019timm}, as well as for a robust ResNet-50 from \citet{bai2021transformers} and a non-robust ResNet-50 from the timm library (shared by \citet{wightman2021resnet}). We scale the perturbations into the $[0, 1]$ range. We show some example images of perturbations in \cref{fig:attacks}. We can see from \cref{tab:perts-classification} that, according to this metric, the perturbations generated for both robust models lead to non-trivial accuracies and that the perturbations generated for XCiT-S12 have indeed semantic characteristics, and more so than those generated for the robust ResNet-50. We provide additional feature visualizations in \cref{sec:gradients-additional}.

\begin{table}[ht]
	\centering
	\caption{\textbf{Semantic characteristics in adversarial perturbations.} For each robust network, we first generate adversarial perturbations ($\eps=\frac{4}{255}$, $\ell_\infty$, scaled to [$0$, $1$]) using untargeted attack. We provide an example visualization of these perturbations in Figure~\ref{fig:attacks}. Measuring the top-5 accuracy using pre-trained ConvNeXt-XL, BeiT-L, and Swin-L models, we observe that robust XCiT-S perturbations indeed have semantic characteristics, even higher than a robust ResNet-50.
	}\label{tab:perts-classification}
	\renewcommand*{\arraystretch}{1.25}
	\resizebox{\linewidth}{!}{
	\begin{tabular}{ccccc}
		\toprule
		\multicolumn{2}{c}{\multirow{2}{*}{Perturbations generator}} & \multicolumn{3}{c}{Classifier}                           \\ \cmidrule{3-5}
		\multicolumn{2}{c}{}                                         & ConvNeXt-XL~\cite{liu2022convnet}                    & BeiT-L~\citep{bao2022beit} & Swin-L~\citep{liu2021swin}         \\ \midrule
		\multirow{2}{*}{Robust}                                      & XCiT-S12~\textit{(ours)}                       & $\mathbf{43.86}$  & $\mathbf{49.52}$  & $\mathbf{40.24}$ \\
		                                                             & GELU ResNet-50~\cite{bai2021transformers}                 & $38.40$  & $45.02$  & $36.70$ \\ \midrule
		\multirow{2}{*}{Non-robust}                                  & XCiT-S12~\cite{elnouby2021xcit}                       & $0.84$   & $0.78$   & $0.84$  \\
		                                                             & ResNet-50~\cite{wightman2021resnet}                      & $0.82$   & $0.74$   & $0.80$  \\
		\bottomrule
	\end{tabular}}
\end{table}

\section{Related work}\label{sec:related-work}

\noindent \textbf{Vision Transformers and variants} The ViT~\cite{dosovitskiy2021image} is an architecture that was adapted for computer vision tasks from the transformer~\cite{vaswani2017attention}, which was first meant for natural language processing. After the introduction of the transformer, several variants have been proposed. Some of these are: DeiT~\cite{touvron2021training}, to reduce the need for pre-training on extremely large datasets, CaiT~\cite{touvron2021going}, to train deeper ViTs, XCiT~\cite{elnouby2021xcit}, to use a more efficient attention-like operation, LeViT~\cite{graham2021levit}, a CNN-ViT hybrid to speed-up transformer inference, and the Swin Transformer~\cite{liu2021swin}, which is a hierarchical transformer which can perform image segmentation and object detection.

\smallskip \noindent \textbf{Robustness of Vision Transformers to non-adversarial perturbations.} Whether ViTs are more robust than CNNs has been a controversial topic so far, with contrasting results based on the different contexts and settings where the models are tested. On the one hand, many recent works~\cite{paul2021vision,bhojanapalli2021understanding,bai2021transformers} agree on the fact that ViTs are more robust than CNNs when it comes to out-of-distribution samples. Although, more recently, \textcite{pinto2022impartial} find that ViTs suffer from simplicity bias similarly to CNNs~\cite{harshai2020pitfalls}, and that, when compared to modern CNNs such as ConvNeXt, there is no clear winner.

\smallskip \noindent \textbf{Adversarial robustness of Vision Transformers} At the same time as they assess ViTs' robustness to natural perturbations, many works also study the robustness of non-adversarially trained ViTs to adversarial examples. In particular, in the case of attacks with $\eps \le 0.01$, ViTs show better robustness than CNNs~\cite{bhojanapalli2021understanding,aldahdooh2021reveal,benz2021adversarial}. However, when running stronger attacks, such as APGD or AutoAttack, with $\eps = \nicefrac{4}{255}$, both ResNets and ViTs have $0\%$ robust accuracy~\cite{shao2021adversarial,bai2021transformers,mahmood2021robustness}. Finally, \textcite{mao2022towards} propose the \emph{Robust ViT} (RVT), a ViT with specific architectural innovations which show additional robustness to FGSM and PGD-5 attacks. However, they do not test the robustness of RVT to stronger attacks such as AutoAttack.

\smallskip \noindent \textbf{Adversarially trained Vision Transformers.} \textcite{shao2021adversarial} adversarially fine-tuned non-robust ViTs, pre-trained on ImageNet-1k. Since they use a non-robust pre-trained network, they achieve suboptimal performance~\cite{hendrycks19using}. \textcite{bai2021transformers} comes closest to our work: they noticed poor performance of DeiT-S transformers with data augmentation in adversarial training. We show that similar challenges persist with other ViT architectures, such as XCiTs. While \textcite{bai2021transformers} advocate for a progressively increasing augmentation budget, we show that avoiding strong data augmentation entirely can achieve state-of-the-art performance. Even further, we perform an in-depth analysis of adversarially robust ViTs in both pre-training and finetuning. \textcite{wu2022towards} propose a technique to make adversarial training of transformers more efficient. This mechanism reduces the training time while increasing the robust accuracy over 1-step FGSM. However, since they use the canonical training recipe from DeiT~\cite{touvron2021training}, the robust accuracy of their models could be sub-optimal. Finally, concurrently with this work, \textcite{mao2022easyrobust} published the checkpoints of several ViTs who are adversarially trained and show promising performance. However, the training recipe for their models is not public and there is no paper discussing their results.

\smallskip \noindent \textbf{Robustness through architectural development.} \textcite{xie2020smooth} observe that, in ResNets and other CNN variants, the usage of smooth activation functions, such as SiLU~\cite{elfwing2018sigmoid} and GELU~\cite{hendrycks2016gaussian} increases the robustness of adversarially trained models: their hypothesis is that smoother activation functions make the inner maximization problem in \cref{eq:adv-train} easier to solve. This behavior is also observed by \textcite{bai2021transformers}: ResNet-50 is more robust when employing GELU. \textcite{huang2021exploring} observe that increasing the capacity of a model helps with robustness, but there is a trade-off given by the fact that additional capacity makes the model less smooth and less robust. However, reducing the capacity of the last stage can improve this trade-off. Similar observations are done in concurrent work by \textcite{wu2021wider} who propose a technique to efficiently find a suitable TRADES' $\lambda$ when training wide models.

\smallskip \noindent \textbf{Improving the robustness-accuracy-efficiency trade-off.} Apart from the work by \textcite{wu2022towards} mentioned above, other previous work addresses the problem of efficiency of robustness in different ways. One line of work focuses on making adversarial training more efficient: \citet{shafahi2019adversarial} proposes to re-use adversarial examples from previous epochs, while \citet{wong2020fast} find an effective way to perform adversarial training with just 1-step FGSM.\@ On the other hand, \citet{sehwag2020hydra,kundu2020tunable} work on compressing the model's size by pruning in ways that are fully compatible with adversarial training. Other works show that both robustness and accuracy are improved when we use either extra unlabeled data~\cite{carmon2019unlabeled} or synthetic data~\cite{sehwag2021robust,gowal2021improving}.
\section{Discussion and Conclusion}\label{sec:conclusion}

\smallskip \noindent \textbf{Architectures and custom recipes.} This work shows that, by shifting architecture, we can significantly improve the robustness of image classification models, by keeping a good accuracy-robustness-efficiency tradeoff. We do so by identifying an architecture that has a good fit for adversarial training: the Cross-Covariance Image Transformer (XCiT). We also show that to achieve optimal results, it is important to find a tailored training recipe, which may differ from the canonical training recipe for standard training. Using this custom recipe, we achieve good results which are better than the current state-of-the-art, both in terms of clean and robust accuracy. On the other hand, we have also tested this training recipe on ConvNeXt, a recently proposed modern convolutional architecture, showing that, with our training recipe, it can reach state-of-the-art performance. We leave to future research whether the performance can be further boosted with a custom-tailored recipe for ConvNeXt and the architectural innovations to come (both for training from scratch and fine-tuning).

\smallskip \noindent \textbf{Fine-tuning.} We further successfully show that ViTs can be efficiently robustly fine-tuned, for larger perturbation sizes, to very high accuracy on smaller datasets. As a matter of fact, our tailored training recipe also works for larger perturbations with minimal changes. This enables efficiently fine-tuning these models to other datasets and doing adversarial training on smaller high-resolution datasets. Moreover, given the trends shown in standard training of ViTs and previous work about adversarial training, we believe that XCiT could further benefit from being trained on larger datasets such as ImageNet-21k and then fine-tuned on downstream datasets. We suggest that researchers should consider this option when doing adversarial training for ViT-like models, given that, as we have shown, a model can be fine-tuned efficiently in a few epochs.

\smallskip \noindent \textbf{Analyses.} We show a potential explanation of the improved robustness of XCiT and ConvNeXt compared to ResNet and XCiT trained with the canonical recipe: for former models, the 1-step attack is more effective throughout the whole training procedure. Finally, we analyze the gradients of our robust XCiT and compare the visualizations to a state-of-the-art robust ResNet: we quantify that the perturbations found for XCiT are more semantic than those of ResNet, suggesting that the robust XCiT's perturbations are more aligned with human perception. We believe that further insightful analyses can be carried on, given the different nature of ViT-like models. For this reason, we release the checkpoints of our models trained for different epsilons. We believe that this enables researchers to do further analyses that will improve our understanding of why such a simple recipe is particularly suitable for adversarial training.

\section{Acknowledgements}
We thank Google's TPU Research Cloud (TRC) Program\footnote{\url{https://sites.research.google/trc/about/}}, which provided us with extremely generous computing resources. We also thank Maksym Andriushchenko, Jacopo Teneggi, Florian Tram{\`e}r, Chong Xiang, and Xinyu Tang for their feedback about this work. This work was also supported in part by the National Science Foundation under grants CNS-1553437 and CNS-1704105, the ARL’s Army Artificial Intelligence Innovation Institute (A2I2), the Office of Naval Research Young Investigator Award, the Army Research Office Young Investigator Prize, Schmidt DataX award, Princeton E-ffiliates Award, and Princeton Gordon Y. S. Wu Fellowship.

\printbibliography

\appendix

\subsection{The vision transformer architecture}\label{sec:vision-transformer}

We now cover the basic building blocks of the (vision) transformer architecture: attention and multi-head attention, the MLP block, positional encoding, and tokens embedding.

\subsubsection{Attention} The attention function maps a query vector and a set of key-value vector pairs to an output vector. In particular, the form of attention by \citet{vaswani2017attention}, called \emph{Scaled Dot-Product Attention}, can be formally expressed as
\begin{equation}
	\text{Attention}(\mQ, \mK, \mV) = \text{Softmax}\left( \frac{\mQ\mK^T}{\sqrt{d_k}} \right) \mV,
\end{equation}
where $\mQ$, $\mK$, and $\mV$ represent respectively the set of queries, keys, and values grouped in matrices, and $d_k$ represents the dimension of the queries and the keys, while the values have dimension $d_v$. The product is scaled by $\frac{1}{\sqrt{d_k}}$ to compensate for the fact that the dot products can reach large values in magnitude when $d_k$ is large. Large values would saturate $\text{softmax}$ and make its gradients very small. In practice, in the context of NLP, attention is computed among different words (or parts of words). Given a set of words (e.g., a sentence), it measures how much a word is dependent on another word in the same set. For instance, in the sentence ``This is a paper about Vision Transformers'', ``this'' will attend to ``is'', which will, in turn, attend to ``paper''. The main advantage of attention, when compared to convolutions, is the ability to capture long-distance dependencies between tokens, which is something convolutions fail to do because of the local nature of the convolution operator. A potential advantage of using attention for computer vision tasks is that it can measure how much a portion of an image attends to another one, enabling the possibility of working more at a global level than a local one, as convolutions do. For instance, in an image of a cat, the tail, or the paws, will attend to the cat's head, and vice-versa.

\subsubsection{Multi-Head Attention}\label{par:mhsa} Before passing $\mQ$, $\mK$, and $\mV$ to the Attention function, \citet{vaswani2017attention} linearly project, using learnable matrices, the inputs into vectors with dimension $d_k$, $d_k$, and $d_v$ respectively. Moreover, instead of doing it just once, they do it $h$ times, and each projection is passed to the $\text{Attention}$ function simultaneously, creating the so-called Multi-Head Attention. After the parallel processing, the resulting matrices are concatenated and linearly projected, using, again, a learnable matrix. Parallel processing enables the model to efficiently process information from different representations of the inputs. Formally,
\begin{equation}
	\begin{split}
		\text{MultiHead}(\mQ, \mK, \mV) & = \text{Concat}(\text{head}_1, \dots, \text{head}_h)\mW^O, \\
		\text{where head}_i & = \text{Attention}(\mQ\mW_i^Q, \mK\mW_i^K, \mV\mW_i^V),
	\end{split}
\end{equation}
where $\mW_i^Q \in \mathbb{R}^{d_\text{model} \times d_k}$, $\mW_i^K \in \mathbb{R}^{d_\text{model} \times d_k}$, $\mW_i^V \in \mathbb{R}^{d_\text{model} \times d_v}$, and $\mW^O \in \mathbb{R}^{hd_v \times d_\text{model}}$ are the learnable matrices used to linearly project the inputs and the result of the Attention operation. Finally, we call Self-Attention the special case where $\mK = \mV$, and --analogously-- Multi-Head Self-Attention (MSA) a Multi-Head Attention in the case where $\mK = \mV$.

\subsubsection{The MLP block and the overall Transformer block}
After computing self-attention for the inputs, the result is passed to a fully connected Multi-Layer Perceptron (MLP) block with one hidden layer. Formally, given an input $x$, and learnable weights and biases $\mW_1$, $\vb_1$, $\mW_2$, $\vb_2$, and an activation function $\rho$, the MLP block can be expressed as
\begin{equation}\label{eq:mlp-block}
	\mlp(\vx) = \rho(\vx\mW_1 + \vb_1)\mW_2 + \vb_2.
\end{equation}
The vision transformer uses the Gaussian Error Linear Unit (GELU)~\cite{hendrycks2016gaussian} activation and apply layer normalization to the input of each MSA and MLP block.

To summarize, given an input $\vx_l$ to the $l$-th Vision Transformer block, a multi-head self-attention block MSA, an MLP block, and a layer normalization block LN, the overall Transformer block can be expressed as:
\begin{equation}\label{eq:vit-block}
	\begin{split}
		\vx_l' & = \vx_l + \msa(\layn(\vx_l)) \\
		\vx_{l + 1} & = \vx_l' + \mlp_{\text{GELU}}(\layn(\vx_l')).
	\end{split}
\end{equation}

\subsubsection{Positional encoding}

The reader can observe that attention, per se, considers the input as a set, and not as a sequence. Hence, all information about the position of the inputs is completely lost. For this reason, \citet{vaswani2017attention} add the so called Positional Encodings to the input embeddings before feeding them to the encoder and the decoder. The positional encodings have the same size as the embeddings of the input tokens, i.e., $d_\text{model}$. In the case of the vision transformer, the positional embeddings are learned parameters.

\subsubsection{Input tokenization and positional encoding}\label{par:tokenization}

Considering each pixel as a token and computing attention between every pixel would be computationally unfeasible, as the attention operation has $\mathcal{O}(n^2)$ complexity for both memory and runtime. For this reason, \citet{dosovitskiy2021image} split the image into non-overlapping input patches. The patches are embedded into tokens by reducing, by the size of the patches, the overall number of inputs to the attention operation. The patches are embedded by linearly projecting them to vectors of dimension $\mathbb{R}^d_\text{model}$. Given an image of size $(H, W)$ and patches of size $(P, P)$, the resulting number of patches is $N = HW / P^2$.

\subsubsection{Class token}\label{par:class-token}

After the input tokens are generated, a vector --- called \texttt{[class]} token --- is prepended to the sequence of tokens and processed along with the other tokens. The initial state of this vector is a learnable parameter of the model. The class token is meant to attend to the most relevant parts of an image, e.g., in the case of an image with a cat, the \texttt{[class]} token will attend to the patches, including the cat's head, its tail, and its whiskers. Finally, at the end of the last block, there is the so-called ``MLP Head'': an MLP which takes as input the \texttt{[class]} token resulting from the last attention block, and maps it to the class predicted for the input.

\subsubsection{Performance and variations of Vision Transformers}\label{sec:deit}

ViTs achieve state-of-the-art performance on several datasets, such as ImageNet-1k~\cite{deng2009imagenet}, CIFAR-10, and CIFAR-100~\cite{krizhevsky09learning}. In particular, their maximum potential is reached when they are pre-trained on larger datasets, such as ImageNet-21k and JFT-300M~\cite{sun2017revisiting}. In this way, they can learn representations that are more generalizable and do not overfit when they are trained on smaller datasets such as CIFAR-10 and CIFAR-100. To reduce the need for pre-training on larger datasets, concurrent work \citet{touvron2021training,steiner2021train} shows that a tuned training recipe, strong regularization, and data augmentations, such as CutMix~\cite{yun2019cutmix}, RandAugment~\cite{cubuk2020randaugment}, MixUp~\cite{zhang2017mixup}, and RandomErasing~\cite{zhong2020random}, lead to a ten-fold decrease in the need of data to achieve the same performance. In particular, \citet{touvron2021training} call the model trained with this training recipe \emph{DeiT}. On the architectural point of view, instead, it has been proposed to specialize layers for patches processing and classification separately, with the aim of more effectively train deeper ViTs~\citep{touvron2021going}, and to use different operations than multi-head self-attention, such as cross-covariance attention~\citep{elnouby2021xcit} or average pooling~\cite{yu2021metaformer}.

\subsection{Transformer variations}\label{sec:transformer-arch}

We now give an overview of two transformer architectures we employ in our experiments, i.e., CaiT~\citep{touvron2021going} and XCiT~\citep{elnouby2021xcit}.

\subsubsection{Class Attention in Transformers (CaiT)}\label{sec:cait}

In order to train deeper ViTs, \citet{touvron2021going} introduce two innovations: \emph{LayerScale} and \emph{Class Attention}.

\paragraph{LayerScale}\label{par:layerscale}
LayerScale consists of two learnable diagonal matrices: one is multiplied to the result of the attention operation, and the other is multiplied to the result of the MLP block. Formally, given the two LayerScale diagonal matrices $\diag(\lambda_{l, 1}, \dots \lambda_{l, d})$ and $\diag(\lambda_{l, 1}', \dots \lambda_{l, d}')$, the transformer block function becomes
\begin{equation}\label{eq:layer-scale}
	\begin{split}
		\vx_l' & = \vx_l + \diag(\lambda_{l, 1}, \dots \lambda_{l, d}) \msa(\layn(\vx_l)) \\
		\vx_{l + 1} & = \vx_l' + \diag(\lambda_{l, 1}', \dots \lambda_{l, d}') \mlp(\layn(\vx_l')).
	\end{split}
\end{equation}
In the case of deeper architectures (i.e., with a total of 24 transformer blocks), these diagonal matrices are initialized to a value $\eps$. This value is equal to $0.1$ for architectures with up to depth 18, $10^{-5}$ for those with depth 24, and $10^{-6}$ for those with depth 38.

\paragraph{Class Attention}
On the other hand, class attention introduces a new way to handle the \texttt{[class]} token: instead of prepending it to the sequence of the input tokens at the beginning of the sequence of transformer blocks, \citet{touvron2021going} first process the input tokens through a series of Transformer blocks (according to \cref{eq:layer-scale}), in a stage called \emph{self-attention} stage. They then prepend the \texttt{[class]} token, and go through a series of blocks composed of a Multi-Head Class Attention block followed by an MLP block. They call this stage \emph{class-attention} stage. A Multi-Head Class Attention block is like a Multi-Head Self-Attention block where only the Attention of the \texttt{[class]} to the other tokens is computed, and the other tokens are left untouched. Formally, given a \texttt{[class]} token $\vx_\text{class}$, a vector $\vz = [\vx_\text{class}; \vx_\text{patches}]$ given by the concatenation of the class token and the patches, learnable weight matrices $\mW_q$, $\mW_k$, $\mW_v$, and $\mW_o$ in $\mathbb{R}^{d \times d}$, and corresponding bias vectors $\vb_q$, $\vb_k$, $\vb_v$, and $\vb_o$ in $\mathbb{R}^d$ where $d$ is the size of the token embeddings, they first perform the projections:
\begin{equation}
	\begin{split}
		\mQ & = \mW_q \vx_\text{class} + \vb_q \\
		\mK & = \mW_k \vz + \vb_k \\
		\mV & = \mW_v \vz + \vb_v.
	\end{split}
\end{equation}
They then compute class attention as
\begin{equation}
	\text{CA}(\mQ, \mK, \mV) = \mW_o \text{Softmax}\left(\frac{\mQ \mK^T}{\sqrt{\nicefrac{d}{h}}}\right) \mV + \vb_o,
\end{equation}
where $\mQ \mK^T \in \mathbb{R}^{h \times 1 \times p}$, and $h$ is the number of heads and $p$ is the number of patches. The class-attention stage is composed of two Class Attention blocks, and the resulting architecture is dubbed \emph{Class Attention in Transformers} (CaiT).

\subsubsection{Cross-Covariance Attention and XCiT}\label{sec:xcit}

\citet{dosovitskiy2021image} show that using a smaller patch size brings better results. However, the attention operation has $\mathcal{O}(n^2)$ complexity for both memory and runtime, making decreasing the patch size hard. For this reason, \citet{elnouby2021xcit} propose an alternative to the attention operation, called \emph{Cross-Covariance Attention}, with complexity $\mathcal{O}(n)$. The corresponding ViT-like architecture is called \emph{Cross-Covariance Image Transformer} (XCiT). Overall, XCiT has a structure analogous to that of CaiT (i.e., with self-attention and class-attention phases), with the difference that it employs Cross-Covariance Attention instead of Self-Attention. As a further difference, models with depth 12 initialize LayerScale's $\eps$ to $1$ instead of $0.1$ (as discussed in Paragraph \ref{par:layerscale}).

\paragraph{Cross-Covariance Attention}
Cross-Covariance Attention (XCA) is an attention mechanism based on cross-covariance, which works along the features dimension, i.e., along each dimension of the token embeddings. Given a queries matrix $\mQ$, a keys matrix $\mK$, and a values matrix $\mV$, cross-covariance attention is defined as
\begin{equation}
	\text{XC-Attention}(\mQ, \mK, \mV) = \mV \text{Softmax}\left( \frac{\hat{\mK}^T \hat{\mQ}}{\tau} \right),
\end{equation}
where $\hat{\mK}$ and $\hat{\mQ}$ are the $\normltwo$-normalized versions (i.e., with unit $\normltwo$ norm) of $\mK$ and $\mQ$. It is called cross-covariance attention as, in the case of self-attention $\hat{\mK}^T \hat{\mQ} = W_k^T X^T X W_q$ is the cross covariance matrix of $\hat{\mK}$ and $\hat{\mQ}$, $\Cov(\hat{\mK}, \hat{\mQ})$. Cross-covariance is linear in time in the number of elements in $X$, i.e., the number of patches $N$. We can interpret XCA as a dynamic, data-dependent, $1 \times 1$ convolution along the axis of the features of the embeddings, as each patch is multiplied by the same data-dependent weight-matrix.

Finally, $\tau$ corresponds to a learnable temperature scaling parameter, which is applied to help the convergence of the training procedure.

\paragraph{Local Patch Interaction}
Given the nature of XCA, the patches do not explicitly interact with each other. For this reason, after computing XCA, \citet{elnouby2021xcit} apply the so-called \emph{Local Patch Interaction} (LPI), which consists of two $3 \times 3$ depth-wise convolutional layers with batch normalization and GELU activation between the two layers.

\paragraph{Convolutional Patch Projection}\label{sec:conv-patch-projection} Differently than the previous works about ViTs introduced above, following \citet{graham2021levit}, \citet{elnouby2021xcit} embed the input patches into tokens using a series of $3 \times 3$ convolutions of stride 2 with GELU activation in between. As an example, for a model with embedding dimension $d_\text{model}$ with patch size 16, an RGB input image of size $(3, 224, 224)$ goes through the following transformations: $(3, 224, 224) \rightarrow (d_\text{model} / 8, 112, 112) \rightarrow (d_\text{model} / 4, 56, 56) \rightarrow (d_\text{model} / 2, 28, 28) \rightarrow (d_\text{model}, 14, 14)$. We note that $224 / 16 = 14$, i.e., the final result, as expected, is a set of $14 \times 14$ vectors of size $d_\text{model}$, each of which is mapped from a patch. Finally, they use fixed, sinusoidal positional encoding as in the original work from \citet{vaswani2017attention}.

\subsection{Data augmentations}\label{sec:data-augmentation}

Apart from classic data augmentation strategies, such as random flipping, there are more advanced data augmentation techniques. The ones employed by \citet{touvron2021training} are \emph{MixUp}, \emph{CutMix}, \emph{RandAugment}, and \emph{Random Erasing}.

\begin{figure}[ht]
	\centering
	\begin{subfigure}{0.49\textwidth}
		\centering
		\includegraphics[width=\textwidth]{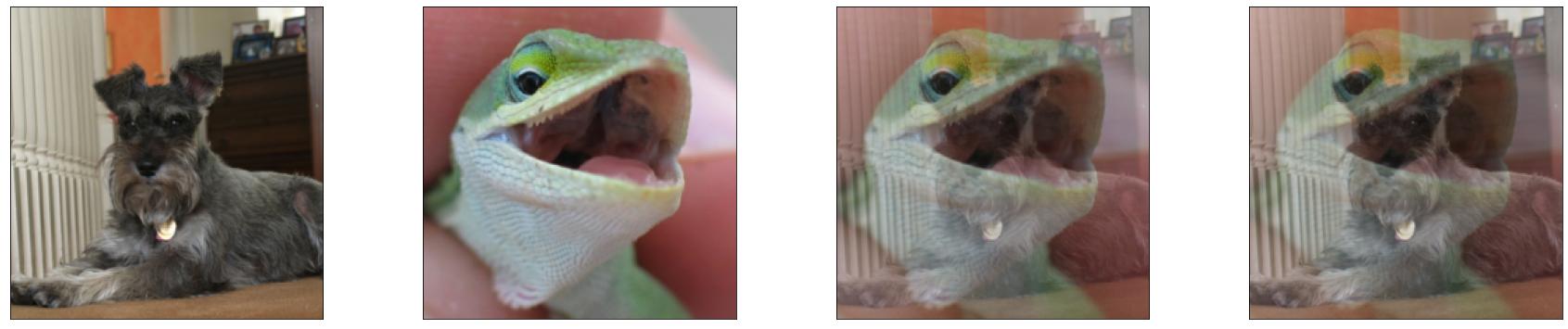}
		\caption{MixUp example.}
	\end{subfigure}
	\begin{subfigure}{0.49\textwidth}
		\centering
		\includegraphics[width=\textwidth]{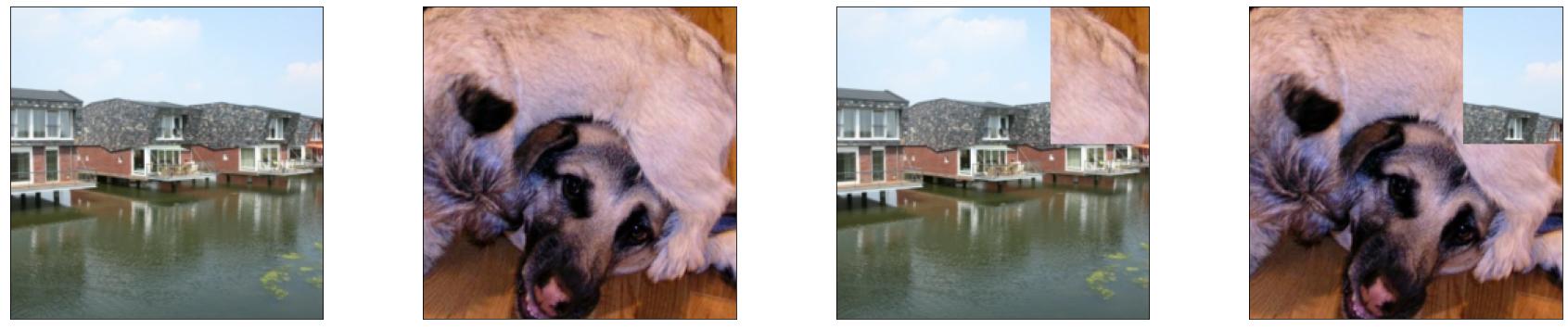}
		\caption{CutMix example.}
	\end{subfigure}
	\caption{Examples for MixUp \figtop{} and CutMix \figbottom{} data augmentations.}\label{fig:cutmix-mixup}
\end{figure}

\smallskip \noindent \textbf{MixUp and CutMix.} MixUp~\cite{zhang2017mixup} consists of creating an image $\tilde{\mX} \in \{0, 1 \}^{H \times W}$ of size (H, W) and a corresponding label $\tilde{\vy}$ as the convex combination of two images and their respective labels. This means that, given two images $\mX_1$ and $\mX_2$, with respective one-hot-encoded labels $\vy_1$ and $\vy_2$, MixUp generates an image $\tilde{\mX} = \lambda \mX_1 + (1 - \lambda) \mX_2$, and the same is applied to the one-hot-encoded labels: the resulting label is $\tilde{\vy} = \lambda \vy_1 + (1 - \lambda) \vy_2$. CutMix~\cite{yun2019cutmix} follows a similar principle by cutting a portion of an image and superimposing it on another image. Formally, the resulting image is computed as $\tilde{\mX} = \mM \odot \mX_1 + (\vone - \mM) \odot \mX_2$, where $\vone$ is the matrix of all ones, and $\mM$ is a masking matrix. In particular, the masking matrix $\mM$ has zeros everywhere apart from the bounding box $\mB$ delimited by the coordinates $(r_x, r_y , r_h, r_w)$, where $r_x$ and $r_y$ are sampled uniformly along the height and the width of the image, $r_h = H \sqrt{1 - \lambda}$ and $r_w = W \sqrt{1 - \lambda}$. In this way, the box is placed randomly in the image and has area proportional to $\lambda$. We show some examples for these data augmentations in \cref{fig:cutmix-mixup}.

\begin{figure*}[ht]
	\centering
	\includegraphics[width=\textwidth]{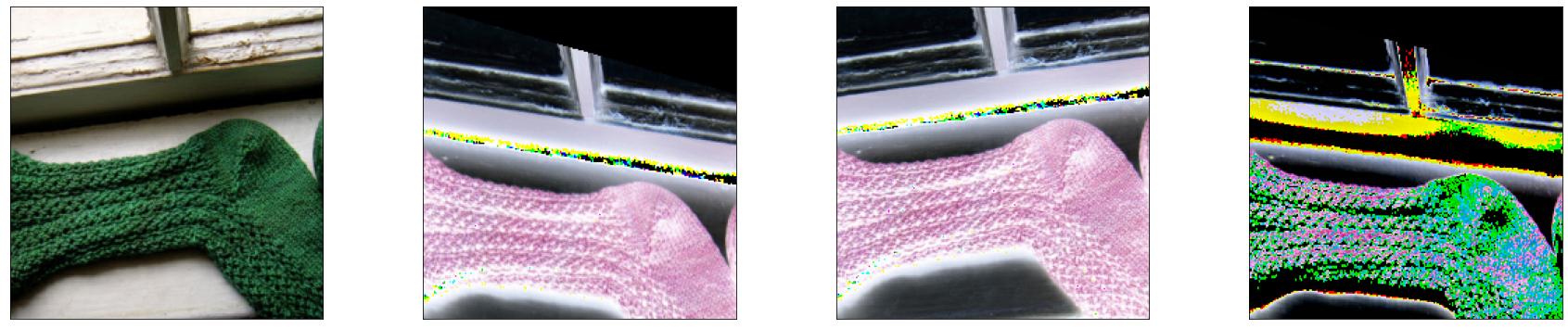}
	\caption{Original image \figleft{} and three examples of augmented images generated by RandAugment}\label{fig:randaugment}
\end{figure*}

\smallskip \noindent \textbf{RandAugment.} RandAugment~\cite{cubuk2020randaugment} improves the so-called \emph{automated augmentations}, which automatically select the best augmentations for a given model and task, among a given list of possible transformations (e.g., rotation and brightness change). Automated augmentations are effective, but need a separate search phase. RandAugment reduces the search space, which enables training without a prior search phase. In particular, given $K$ augmentations, RandAugment chooses each transformation with probability $\nicefrac{1}{K}$. We show an example for three RandAugment augmentations in \cref{fig:randaugment}.

\begin{figure}[!ht]
	\centering
	\includegraphics[width=0.49\textwidth]{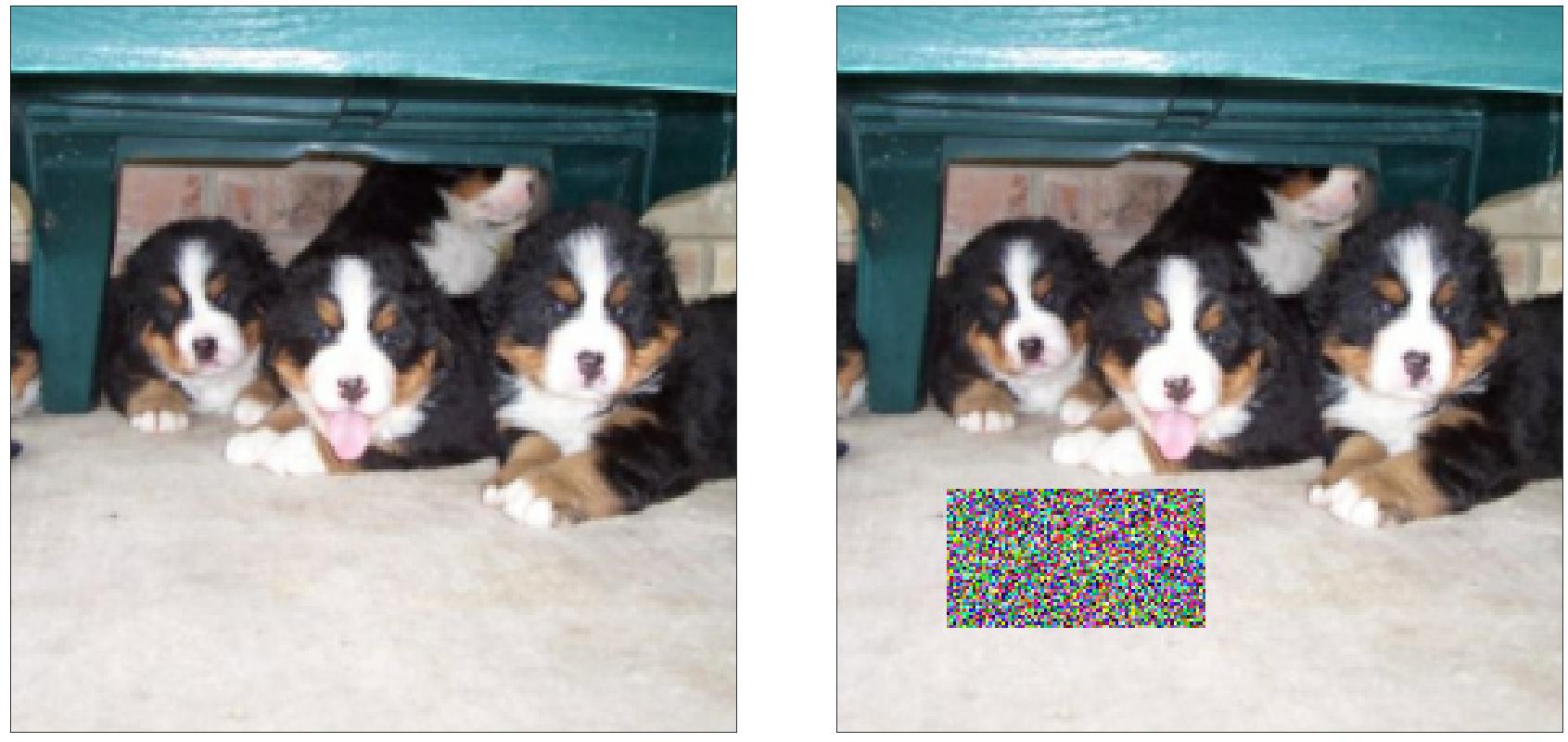}
	\caption{Original image \figleft{} and example of erasure generated by Random Erasing \figright{}.}\label{fig:random-erasing}
\end{figure}

\smallskip \noindent \textbf{Random Erasing.} Finally, Random Erasing~\cite{zhong2020random} randomly selects a portion of pixels in an image, and occludes them, either by setting them to 0 or by sampling their value from a normal distribution with mean and standard deviation equal to those of the dataset. We show an example for Random Erasing in \cref{fig:random-erasing}.

\subsection{Setup and hyperparameters}\label{sec:setup}

In this section, we give additional details about the training, evaluation, and implementation setups.

\smallskip \noindent \textbf{Training hyperparameters.} Apart from the experiments on larger models, we run all the training runs using VMs with 8 TPUv3 cores. In all the runs, unless otherwise stated, we use the same setup as the one used for DeiT~\citep{touvron2021training}. We use as a batch size $64 \times 8 = 512$ (i.e. $64$ samples per TPU core), and the learning rate is chosen according to the formula provided by \citet{touvron2021training} (i.e. $\text{lr} = 0.0005 \times \frac{\text{batch size}}{512}$), which corresponds to $0.0005$ with the batch size we use in most experiments. For all the training runs on ImageNet-1k, we train the model for 110 epochs, with a learning rate cosine decay with a final value of $5 \times 10^{-5}$, a 10-epochs warm-up from $5\times 10^{-6}$ and 10 epochs cool-down. We use the AdamW optimizer~\cite{loshchilov2017decoupled}. Apart from the architecture ablation (\cref{sec:arch-ablation}), we do not employ \emph{repeated augmentations}~\citep{hoffer2020augment} to save training time. Repeated augmentations consist of repeating each batch $3$ times: the first one without data augmentations and the following ones with it. Hence, employing repeated augmentations would be equivalent to doing 3$\times$ the number of epochs. On the other hand, for XCiT-M12, XCiT-L12, ConvNeXt, and PoolFormer, we use TPUv4 pods with either 32 or 64 TPU cores. As mentioned above, we increase the total batch size according to the model size and on the number of devices in use, and we scale the learning rate according to the rule stated above.

\smallskip \noindent \textbf{Training runtime.} We train all three XCiT variants on a pod with 64 TPUv4 cores to compare the training time. We use the largest batch size that can fit into each device, which is 256 for XCiT-S12 and XCiT-M12, and 128 for XCiT-L12. We scale the learning rates as described in the paragraph above. The total training time for XCiT-S12 is 19h30m, for XCiT-M12 it is 33h, and for XCiT-L12 it is 39h.

\smallskip \noindent \textbf{Training attack setup.} Finally, unless otherwise stated, we use FGSM for adversarial training~\citep{wong2020fast}, initializing the adversarial perturbation to be uniformly distributed in $[-\eps, \eps]$, and adding $10^{-5}$ to avoid numerical instability. Moreover, we apply early-stopping, i.e., we evaluate the checkpoint at the epoch where the model was performing best in terms of FGSM accuracy on the test set.

\smallskip \noindent \textbf{Large epsilon pre-training setup.} We pre-train an XCiT-S12 on ImageNet-1k with $\eps = \nicefrac{8}{255}$. Using the same setup as the $\eps = \nicefrac{4}{255}$ training, we observe strong label leaking (which was also observed by previous work on ViTs adversarial training~\citep{herrmann2021pyramid}). To solve this, we use 2-steps FGSM instead of 1-step FGSM as the attack for adversarial training. By doing so, we manage to train a model which has 25.00\% AutoAttack accuracy and 63.46\% clean accuracy on the subset of 5000 images from RobustBench when using $\eps = \nicefrac{8}{255}$ as attack budget. We similarly train an XCiT-M12 and an XCiT-L12, using the same setup. Moreover, as a baseline, we attempt to pre-train an XCiT-S12 robust to $\eps = \nicefrac{8}{255}$ perturbations using the canonical training recipe. However, we observe that the training fails. For this reason, we adopt the epsilon warm-up from our tailored training recipe. Finally, we pre-train a ResNet-50 with GELU activation function, using the same setup as \citet{bai2021transformers}. To validate the correctness of our implementation and setup, we first successfully reproduce their results with $\eps = \nicefrac{4}{255}$. However, when training a model robust to $\eps = \nicefrac{8}{255}$ perturbations with this setup, the resulting model is worse than the ReLU ResNet-50 pre-trained by \citet{salman2020adversarially}\footnote{The checkpoints from this paper can be downloaded from \url{https://github.com/microsoft/robust-models-transfer}}. For this reason, in the fine-tuning experiments, we fine-tune this network. We show the full pre-training results in \cref{tab:pre-training}.

\smallskip \noindent \textbf{High-resolution finetuning setup.} We fine-tune the model pre-trained on ImageNet-1k with $\eps = \nicefrac{8}{255}$ on the high-resolution datasets Caltech-101 and Oxford Flowers. We fine-tune using $\eps = \nicefrac{8}{255}$ for 20 epochs and the same training recipe as the one used for pre-training, with the difference that we do adversarial training with 1-step FGSM instead of 2-steps, and we do not employ a warm-up for $\eps$. We also do a fine-tuning run without adversarial training to better quantify the clean-robust accuracy tradeoff.

\smallskip \noindent \textbf{Low-resolution finetuning setup.} The pre-trained XCiT models are meant for inputs with patch size 16. However, such a patch size would be too large for datasets of smaller images such as CIFAR-10 and CIFAR-100 (which have 32$\times$32 resolution). For this reason, we need a way to adapt the model to support a different patch size. Previous work~\citep{shao2021adversarial} achieves this by down-sampling the weights of the convolutional layer that is used by ViT to embed each patch. However, XCiT uses 4 subsequent convolutional layers to embed 16$\times$16 patches into 1-D vectors, and each layer has stride 2~\citep{graham2021levit}. To embed 4$\times$4 patches, we need to use just 2 subsequent convolutions with stride 2. For this reason, we adapt our model by setting the stride of the first two convolutional layers to 1. Regarding the ResNet-50, instead, we substitute the first convolutional layer (which has originally kernel size 7 and stride 2) with one with kernel size 3 and stride 1, and we remove the first pooling layer. Given the smaller size and resolution of the CIFAR datasets, we fine-tune the robustly trained model using TRADES~\citep{zhang2019theoretically}, with PGD-10 as the attack. Similar to the high-resolution datasets, we fine-tune for 20 epochs. However, we remove the color jitter data augmentation, as the inputs are smaller and would make the task too hard. Moreover, we search for the best learning rate, which we find to be $2 \times 10^{-4}$ (as opposed to $5.0 \times 10^{-5}$ that we used for pre-training and the high-resolution datasets). We probably need a larger learning rate to better tune the input embedding layer whose structure we change. Finally, given the smaller resolution of the images, we change the values for the random scale and crop data augmentation as follows: the ratio of possible crop ranges from $[0.75, 1.33]$ to $[0.95, 1.05]$, and the input re-scaling range from $[0.08, 1.0]$ to  $[0.8, 1.2]$. If we kept these large ranges, then very few pixels of the original image would remain after cropping and resizing, hence the task would be too hard, and the model would underfit.

\smallskip \noindent \textbf{Evaluation setup.} For the final ablation and the scaled-up models trained on ImageNet-1k, we run AutoAttack~\citep{croce2020reliable}, an ensemble of white- and black-box attacks. We run the attack on the subset of 5000 ImageNet-1k images used for the RobustBench benchmark~\citep{croce2020robustbench}. Given that AutoAttack is composed of four attacks, two of which are black-box, it is computational expensive. To strike a balance between strength and computational cost, instead, we assess the robustness of the individual ablations using APGD-CE~\citep{croce2020reliable}. APGD-CE is a parameter-free attack, which is the first of the ensemble that makes up AutoAttack. We run this attack with 5 restarts and 100 iterations, the same settings of the attack that is part of AutoAttack. Finally, we compute the FLOPs and number of parameters using the \texttt{fvcore} library\footnote{https://github.com/facebookresearch/fvcore}.

\smallskip \noindent \textbf{Additional implementation details.} We base our implementation on the PyTorch Image Models repository~\citep{rw2019timm}\footnote{\url{https://github.com/rwightman/pytorch-image-models/}}, which includes the \texttt{timm} library and provides a template training script. This library uses the PyTorch framework~\citep{paszke2017automatic}. In particular, given that we run our experiments on Tensor Processing Unit (TPU) devices, we use the PyTorch XLA library, which compiles PyTorch code to the XLA\footnote{XLA stands for Accelerated Linear Algebra, more information about the XLA compiler can be found here: \url{https://www.tensorflow.org/xla/architecture}} Intermediate Representation (XLA IR), needed to run the computations on TPUs. The XLA IR consists of a graph representing the computation performed on tensors. The graph is then compiled and optimized (e.g., by fusing operations when possible). To use \texttt{timm}'s utility functions to work with TPUs, we use the \texttt{bits\_and\_tpu} branch of PyTorch Image Models\footnote{\url{https://github.com/rwightman/pytorch-image-models/tree/bits_and_tpu}. The branch may be eventually merged into \texttt{main}. Hence, this URL may become invalid.}, which introduces the compatibility of the library with XLA and TPUs. We adapt \texttt{timm}'s default training script to perform adversarial training.

Because of an existing bug we identified in PyTorch XLA\footnote{\url{https://github.com/pytorch/xla/issues/3361}}, when we run the attacks (e.g., FGSM) during training, we have to set the model to the \texttt{.train()} mode (which influences the behavior of batch normalization layers). However, this should not impact the overall robustness of the models~\citep{pang2020bag}. Moreover, while all the AutoAttack and APGD-CE evaluations are run on V100 GPUs, hence with the model in \texttt{.eval()} mode, in the case of the attack effectiveness experiment (sec:attack-effectiveness) we run the evaluations on TPUs, thus with the model in \texttt{.train()} mode. We do so as the large number of evaluations we run (160) would have been unfeasible with our GPU compute budget.

\smallskip \noindent \textbf{Carbon emissions.} Finally, the carbon footprint of the project, measured via Google Cloud's Carbon Footprint Console\footnote{\url{https://cloud.google.com/carbon-footprint}}, is 33 $\text{kgCO}_2$. For scale, a flight from Paris to London generates around 55.7 $\text{kgCO}_2$ per person in economy class\footnote{Computed on \url{https://www.icao.int/environmental-protection/Carbonoffset/Pages/default.aspx}}.

\subsection{Additional ablation results}\label{sec:add-results}

We show in \cref{tab:data-aug-full} the full results for the data augmentation ablation. We can observe that the setups with the heaviest data augmentations rank at the bottom.

\begin{table*}[h!]
    \footnotesize
	\renewcommand{\arraystretch}{1.2}
	\centering
	\caption{\textbf{Weak data augmentation is better.} The strategies that perform best are those with just Random Erasing, or no heavy augmentation, such as RandAugment, at all. We report the full results results in this table, sorted by APGD-CE accuracy. In all the runs we keep weak data augmentation that are commonly used (random flip and crop, and color jitter). (Arch: XCiT-N12)}\label{tab:data-aug-full}
	\begin{tabular}{ccccccc} \toprule
		\multicolumn{4}{c}{Data Augmentation Policy} &                                                                                                                                                           & \multicolumn{2}{c}{Accuracy} \\ \midrule
		MixUp                                        & CutMix & RandAugment                                                 & Random Erasing                                                                     & & \textit{Clean} & \textit{APGD-CE} \\ \midrule
		\xmark                                       & \xmark & \xmark                                                      & \cmark                                                                             & & \textbf{67.28} & \textbf{39.22}   \\
		\xmark                                       & \xmark & \xmark                                                      & \xmark                                                                             & & 66.78          & \textbf{39.22}   \\
		\cmark                                       & \xmark & \xmark                                                      & \xmark                                                                             & & 61.04          & 38.56            \\
		\cmark                                       & \xmark & \xmark                                                      & \cmark                                                                             & & 60.46          & 38.26            \\
		\cmark                                       & \cmark & \xmark                                                      & \xmark                                                                             & & 62.04          & 38.18            \\
		\xmark                                       & \xmark & \cmark                                                      & \xmark                                                                             & & 65.34          & 37.64            \\
		\xmark                                       & \xmark & \cmark                                                      & \cmark                                                                             & & 64.76          & 37.62            \\
		\cmark                                       & \cmark & \xmark                                                      & \cmark                                                                             & & 59.80          & 37.20            \\
		\cmark                                       & \xmark & \cmark                                                      & \xmark                                                                             & & 57.16          & 36.74            \\
		\xmark                                       & \cmark & \xmark                                                      & \xmark                                                                             & & 61.62          & 36.30            \\
		\cmark                                       & \cmark & \cmark                                                      & \xmark                                                                             & & 57.60          & 36.06            \\
		\xmark                                       & \cmark & \xmark                                                      & \cmark                                                                             & & 61.70          & 35.74            \\
		\cmark                                       & \xmark & \cmark                                                      & \cmark                                                                             & & 55.64          & 35.70            \\
		\cmark                                       & \cmark & \cmark                                                      & \cmark                                                                             & & 55.96          & 35.38            \\
		\xmark                                       & \cmark & \cmark                                                      & \xmark                                                                             & & 56.64          & 32.92            \\
		\xmark                                       & \cmark & \cmark                                                      & \cmark                                                                             & & 55.64          & 32.40            \\
		\bottomrule
	\end{tabular}
\end{table*}

\subsection{PGD Results}

We report the results for PGD attacks in \cref{tab:pgd-results}.

\begin{table*}[h]
\footnotesize
\renewcommand{\arraystretch}{1.2}
\centering
\caption{\textbf{PGD accuracy decreases with the number of steps.} We run the PGD attack with 5, 10, 50, and 100 steps for $\varepsilon = \nicefrac{8}{255}$. We observe that, as expected, the robust accuracy gently plateaus at 50 steps, with no significant difference between 50 and 100 steps, for all three models. We run the attack on the full ImageNet validation set, using as a step size $1.5 \cdot \nicefrac{\eps}{n})$, where $n$ is the number of attack steps.}
\label{tab:pgd-results}
\begin{tabular}{lccccc}
\toprule
\multirow{2}{*}{Model} & \multirow{2}{*}{Clean Accuracy} & \multicolumn{4}{c}{Robust Accuracy}   \\ \cmidrule{3-6} 
                       &                                 & PGD-5   & PGD-10  & PGD-50  & PGD-100 \\ \midrule
XCiT-S12               & $72.34$                         & $49.16$ & $48.91$ & $48.71$ & $48.69$ \\
XCiT-M12               & $74.04$                         & $51.96$ & $51.71$ & $51.55$ & $51.53$ \\
XCiT-L12               & $73.76$                         & $53.75$ & $53.52$ & $53.37$ & $53.36$ \\ \bottomrule
\end{tabular}
\end{table*}

\subsection{Pre-training results}

We show in \cref{tab:pre-training} the results for training runs with $\eps = \nicefrac{8}{255}$. These are the models we use for fine-tuning.

\begin{table*}[h!]
	\footnotesize
	\renewcommand{\arraystretch}{1.2}
	\centering
	\caption{\textbf{The training recipe also works for larger epsilons.} Results for training with $\eps = \nicefrac{8}{255}$: we can observe that, for XCiT, the performance improves with scale.}\label{tab:pre-training}
	\begin{tabular}{lccccc} \toprule
		Model             & Clean Accuracy & AA Accuracy \\ \midrule
		GELU ResNet-50    & 58.08          & 17.14       \\
        ReLU ResNet-50~\citep{salman2020adversarially}    & 54.90          & 19.72       \\
		\textbf{XCiT-S12} & 63.46          & 25.00       \\
		\textbf{XCiT-M12} & 67.80          & 26.58       \\
		\textbf{XCiT-L12} & 69.24          & 28.74       \\
		\bottomrule
	\end{tabular}
\end{table*}

\subsection{Additional plots regarding attack effectiveness}

\begin{figure*}[h]
    \centering
    \includegraphics[width=0.7\textwidth]{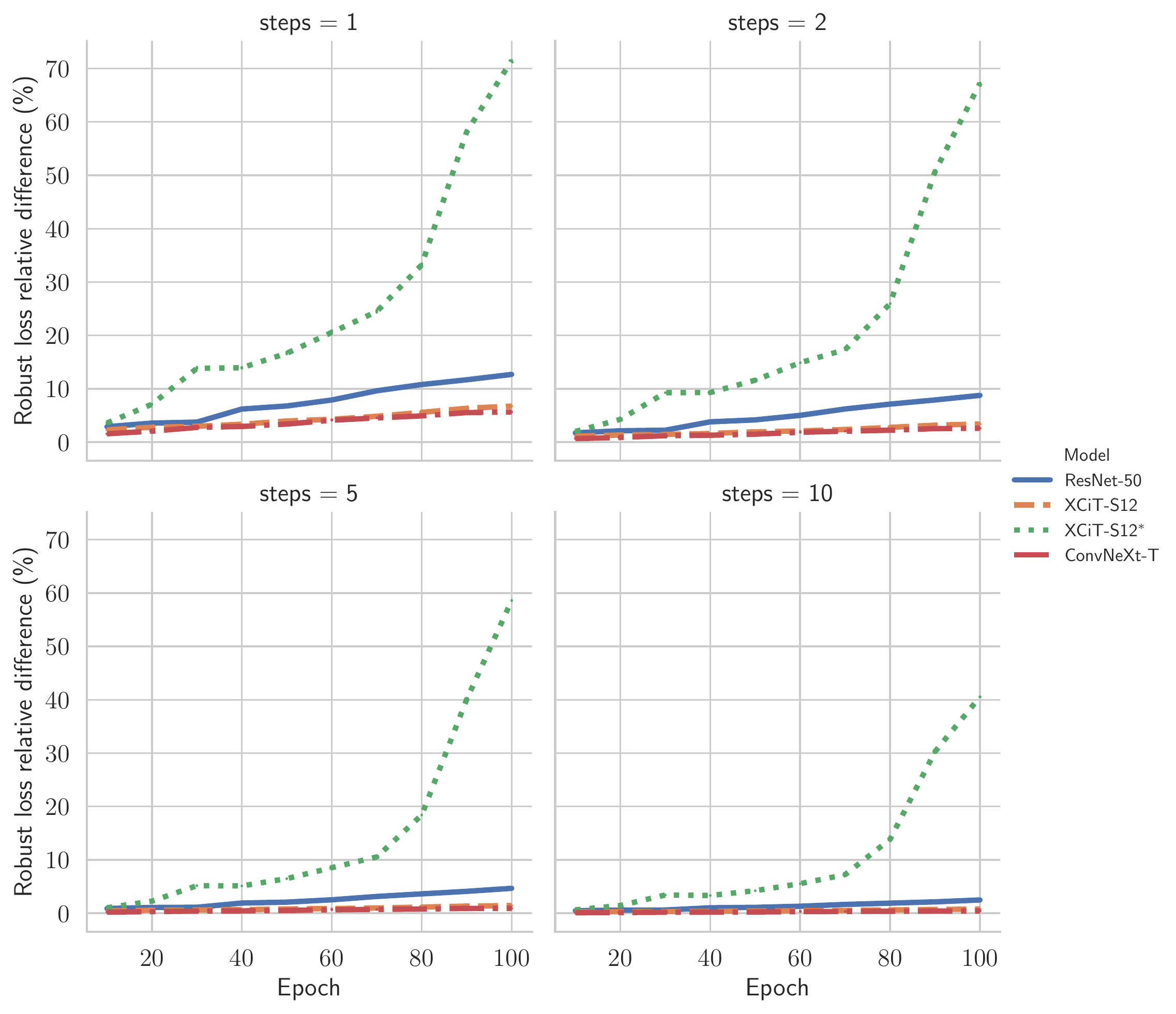}
    \caption{Comparison, across different attack steps, of the relative difference between the adversarial loss computed with the given attack steps and the adversarial loss computed with PGD-200, and how this quantity changes every 10 training epochs.}
    \label{fig:steps_vs_relative_difference}
\end{figure*}

\begin{figure*}[h]
    \centering
    \includegraphics[width=0.5\textwidth]{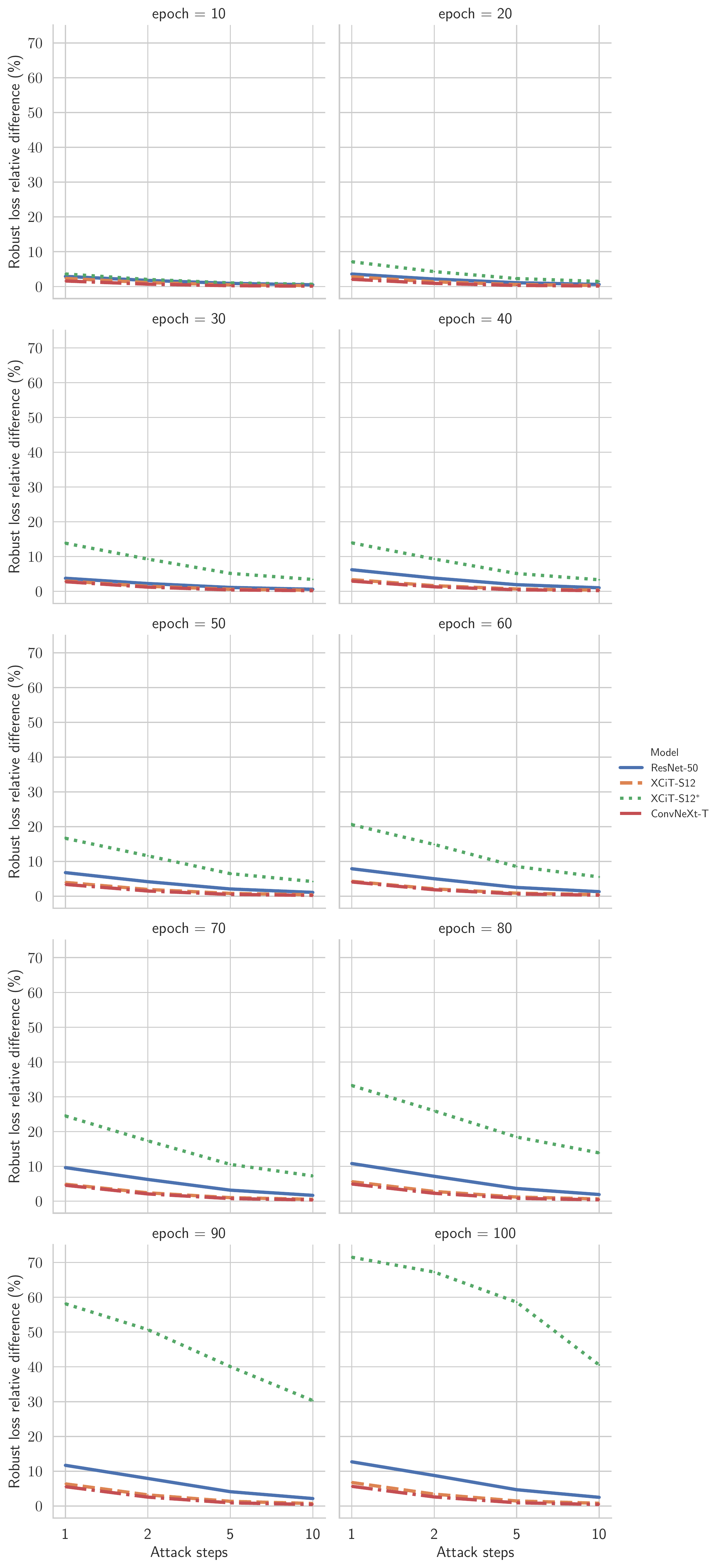}
    \caption{Comparison, every 10 epochs, of the relative difference between the adversarial loss computed with different numbers of attack steps and the adversarial loss computed with PGD-200.}
    \label{fig:steps_unaggregated}
\end{figure*}

We show, in \cref{fig:steps_vs_relative_difference,fig:steps_unaggregated} additional results regarding the attack effectiveness experiment.

\subsection{Additional samples for the optimization sanity checks}\label{sec:additional-sanity-checks}

\begin{figure*}[ht]
	\centering
	\includegraphics[height=0.95\textheight]{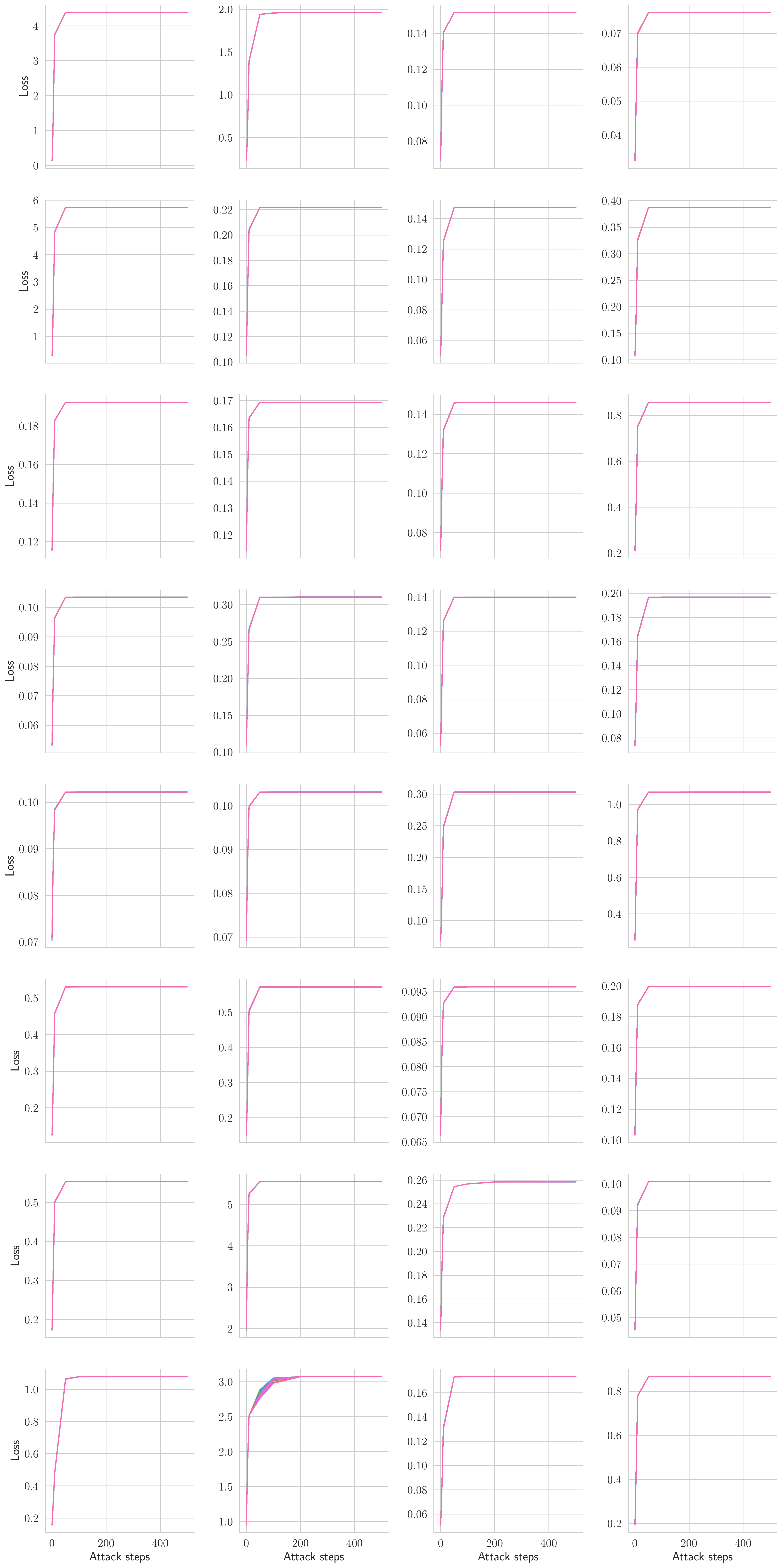}
	\caption{Comparison between different runs of PGD attacks with different numbers of steps for XCiT-S12, for 32 different random points from ImageNet-1k.}\label{fig:attack-convergence-xcit-all}
\end{figure*}

\begin{figure*}[ht]
	\centering
	\includegraphics[height=0.95\textheight]{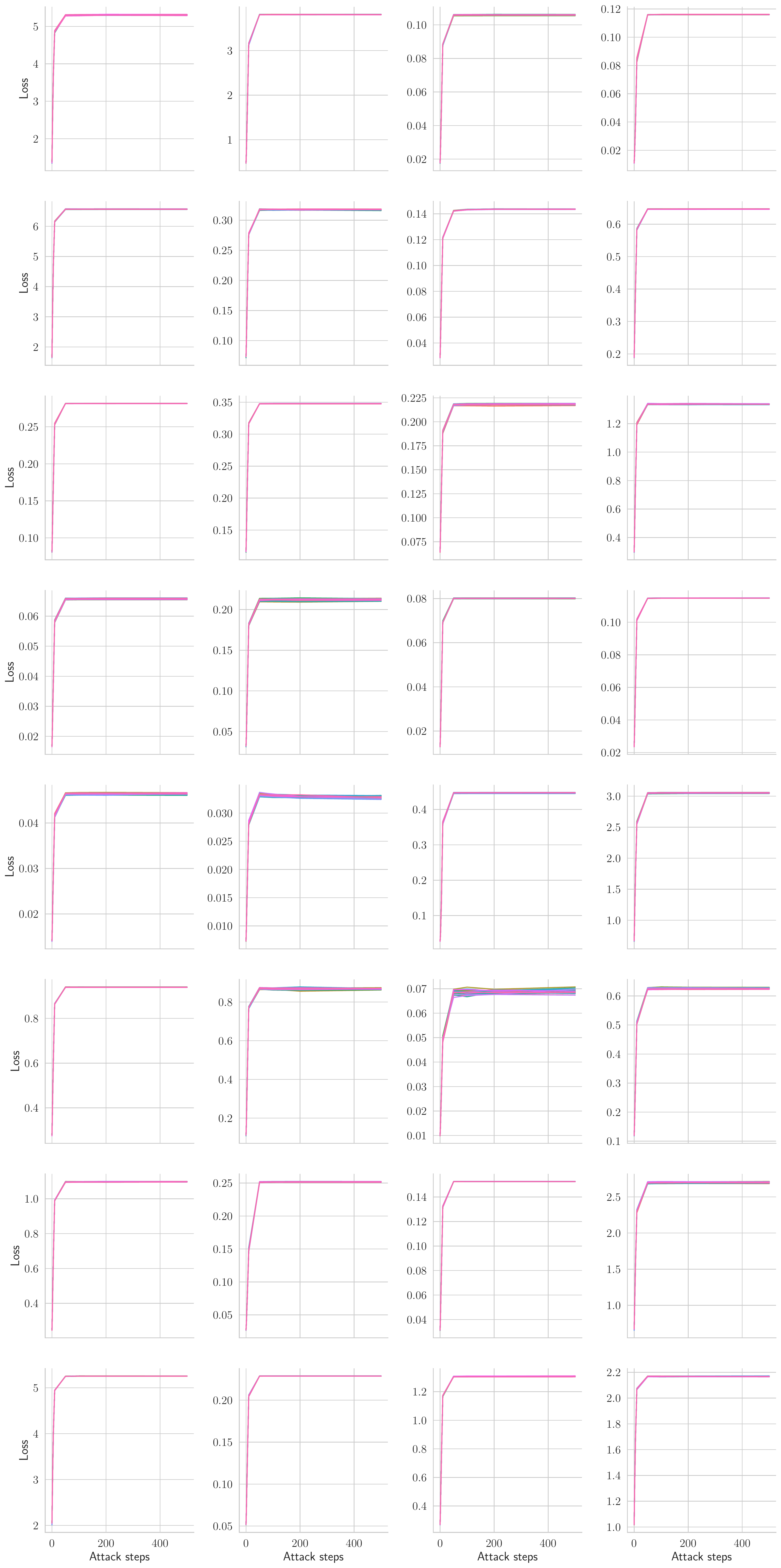}
	\caption{Comparison between different runs of PGD attacks with different numbers of steps for GELU ResNet-50, for 32 different random points from ImageNet-1k.}\label{fig:attack-convergence-resnet-all}
\end{figure*}

\begin{figure*}[ht]
	\centering
	\includegraphics[height=0.95\textheight]{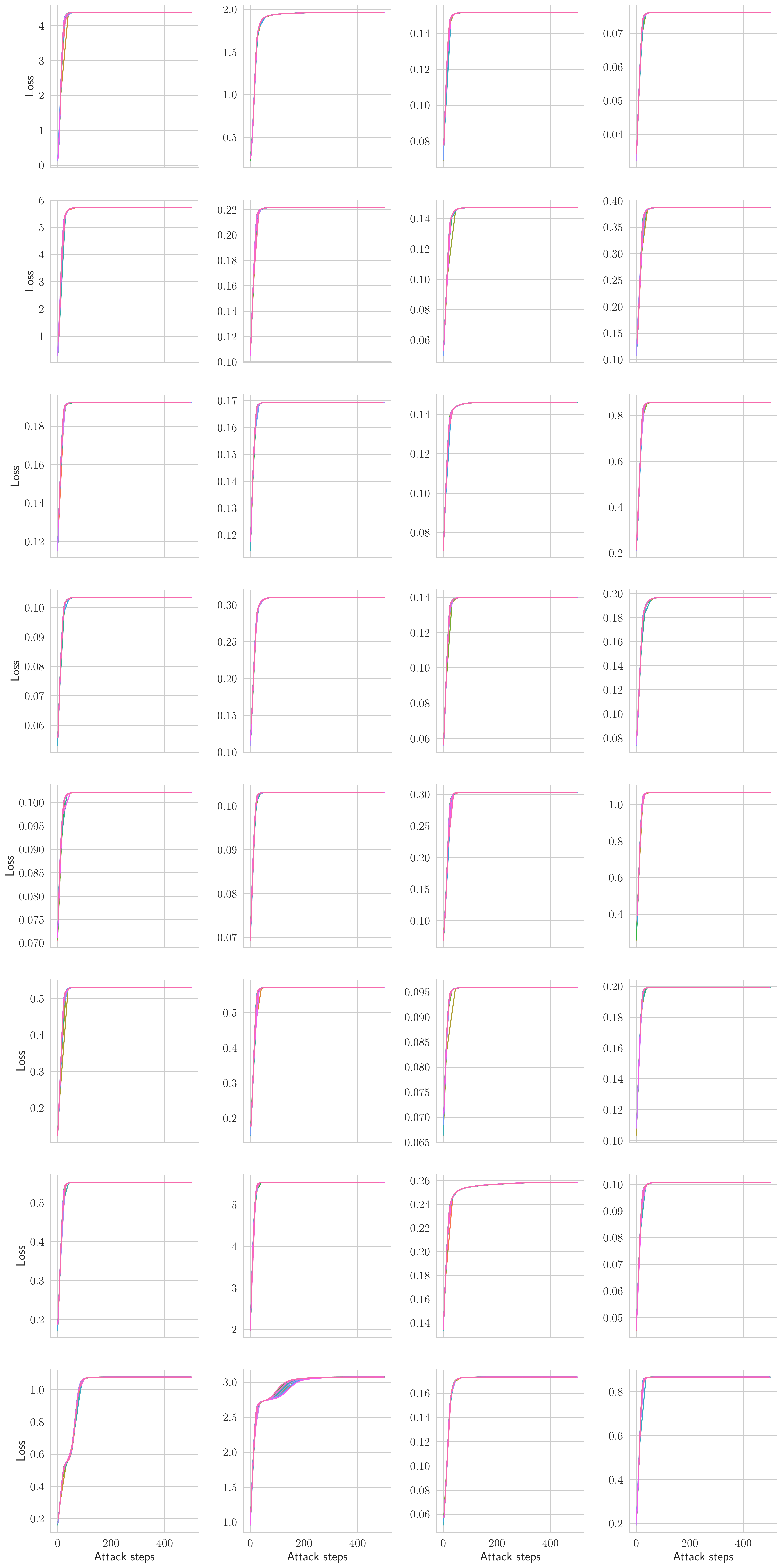}
	\caption{Evolution of the loss for different runs of an attack, using a large step size for XCiT-S12, for 32 different random points from ImageNet-1k.}\label{fig:attack-convergence-one-run-xcit-all}
\end{figure*}

\begin{figure*}[ht]
	\centering
	\includegraphics[height=0.95\textheight]{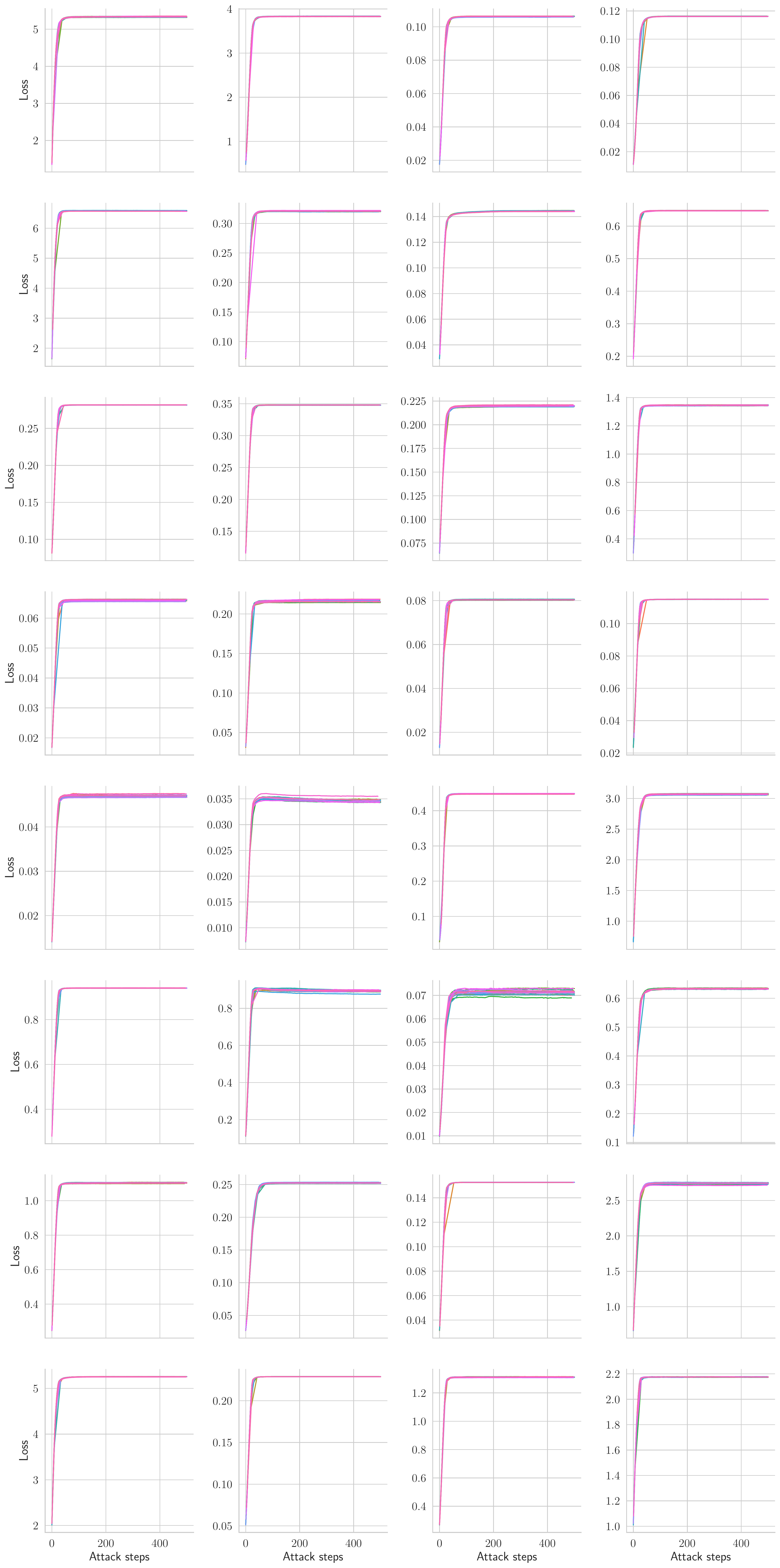}
	\caption{Evolution of the loss for different runs of an attack, using a large step size for GELU ResNet-50, for 32 different random points from ImageNet-1k.}\label{fig:attack-convergence-one-run-resnet-all}
\end{figure*}

We show the loss for different attack steps for 32 different samples from ImageNet-1k in \cref{fig:attack-convergence-xcit-all} (XCiT-S12), and \cref{fig:attack-convergence-resnet-all} (ResNet-50), and the loss progression of the loss in one attack targeting the same 32 samples in \cref{fig:attack-convergence-one-run-xcit-all} (XCiT-S12), and \cref{fig:attack-convergence-one-run-resnet-all} (ResNet-50).

\subsection{Additional experiment about XCiT's gradients}\label{sec:gradients-additional}

\begin{figure*}[ht]
	\centering
	\includegraphics[width=\vizwidht]{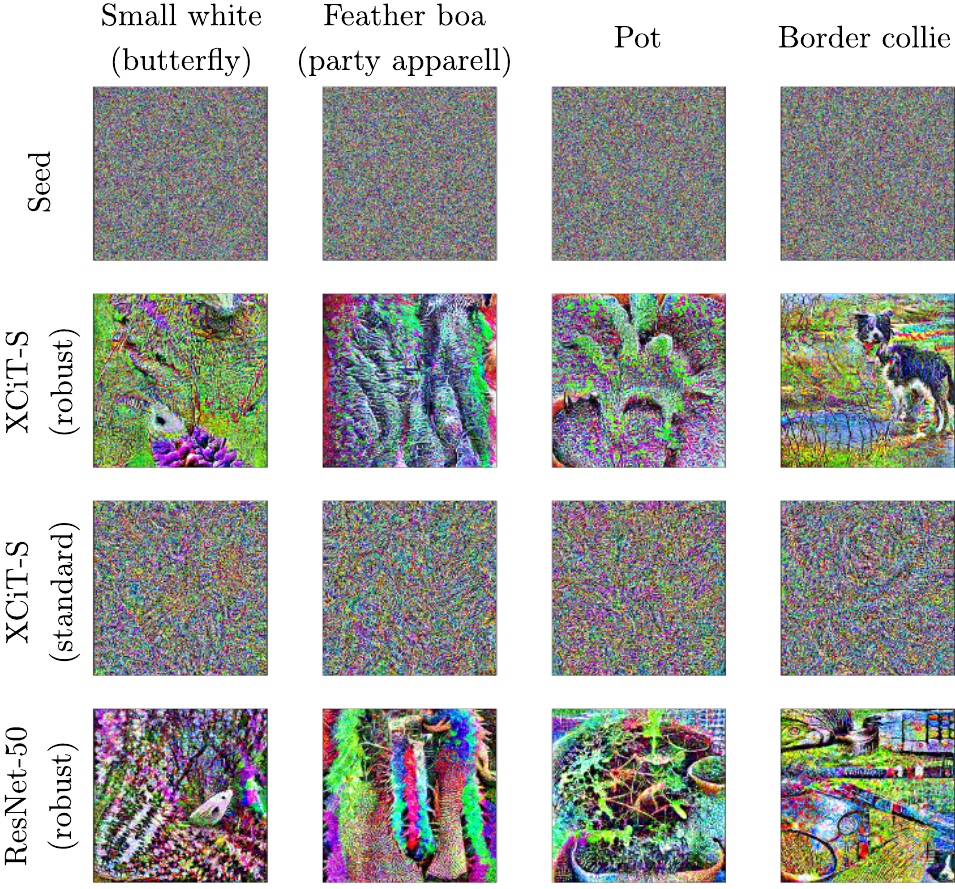}
	\caption{Comparison between the gradient accumulation for a robust XCiT-S12 and a non-robust ResNet-50.}\label{fig:gradients}
\end{figure*}

\smallskip \noindent \textbf{Direct feature visualization.} Gradients of robust CNNs are more aligned with human perception~\citep{engstrom2019adversarial}. In particular, in their work, they introduce a visualization technique called direct feature visualization, by which they maximize the output of a model at a specific activation in the penultimate layer by optimizing an input image via PGD.\@ They observe that the images generated in this way using an adversarially-trained model contain semantically meaningful information without the need for regularization terms on the input. We explore a variation of this experiment: instead of maximizing a specific activation, starting from uniformly random inputs, we run a targeted attack that targets a random class, i.e., we change the input so that it is classified with the given class with the highest confidence. We do so by optimizing the input via PGD-100, using $\eps = 15$. To the best of our knowledge, we are the first to run a similar experiment on a robust ViT-like model. We can see a set of random images and classes in \cref{fig:gradients}.

We make the following observations: 1) The non-robust XCiT gradients have no semantic meaning. 2) Regarding the first image on the left, whose target class is ``small white'' (a butterfly species), for both the robust models, we can see a white portion in the shape of a butterfly with a black spot, which is what a small white butterfly looks. 3) Regarding the image targeting ``feather boa'' (feathery party apparel), we can see, in the case of the robust XCiT-S12, long, colorful structures with feather-like edges. 4) For the images targeting the ``pot'' class, we can see the borders of a pot in the case of the robust XCiT-S12. We can also observe that we can see some plants as well, meaning that, probably, in the datasets, pots are most often represented when containing plants. 5) Finally, in the last image on the right, which should target the ``border collie'' class, we can see a Border Collie for the robust XCiT-S12 and the head of one in the lower right corner for the robust ResNet-50.

\smallskip \noindent \textbf{Perturbations visualization.}
We also visualize the adversarial perturbations generated with a PGD-100 attack for a robust and a non-robust XCiT, compared to those generated for a robust ResNet-50. Given that a perturbation $\vdelta$ is in $[-\eps, \eps]$, we rescale it to $[0, 1]$ to visualize it as an image. For this reason, we compute the visualized images $\delta_\text{viz} = \frac{\delta + \eps}{2 \eps}$ and we visualize the intensity of the perturbation by transforming the image to grey-scale colors.

We can see a random sample of images and their respective perturbations in \cref{fig:attacks}. We note that the shapes of the original images are visible in the robust XCiT and ResNet perturbations, while they are not in the non-robust ones.

\end{document}